\documentclass[lettersize,journal]{IEEEtran}
\usepackage{amsmath,amsfonts}
\usepackage{amssymb}
\usepackage{array}
\usepackage[caption=false,font=footnotesize,labelfont=rm,textfont=rm]{subfig}
\usepackage{textcomp}
\usepackage{stfloats}
\usepackage{url}
\usepackage{verbatim}
\usepackage{multicol}
\usepackage{graphicx}
\usepackage[ruled,linesnumbered]{algorithm2e}
\usepackage{booktabs,multirow,bm}
\usepackage{tabularx}
\usepackage{xcolor}
\usepackage{cite}
\usepackage{makecell}
\usepackage{algpseudocode}
\usepackage{algorithmicx}
\DeclareMathAlphabet\mathbfcal{OMS}{cmsy}{b}{n}
\begin{document}

% 字体首字母大写..?
\title{Is FISHER All You Need in The Multi-AUV Underwater Target Tracking Task?}

\author{Guanwen Xie$^\dagger$, \IEEEmembership{Student Member, IEEE}, Jingzehua Xu$^\dagger$, \IEEEmembership{Student Member, IEEE}, Ziqi Zhang,\\Xiangwang Hou, \IEEEmembership{Member, IEEE}, Dongfang Ma, \IEEEmembership{Member, IEEE}, Shuai Zhang, \IEEEmembership{Member, IEEE},\\Yong Ren, \IEEEmembership{Senior Member, IEEE} and Dusit Niyato, \IEEEmembership{Fellow, IEEE} % ~ for space
        % <-this % stops a space
%\thanks{Preliminary version. Subject to change.}
% <-this % stops a space
\thanks{This work of Xiangwang Hou was supported by the National Natural Science Foundation of China under grant No. 623B2060. This work of Yong Ren  was partly supported by the National Natural Science Foundation of China under Grants No. 62127801 and No. U24A20213. This research is supported by Singapore Ministry of Education (MOE) Tier 1 (RG87/22 and RG24/24), the NTU Centre for Computational Technologies in Finance (NTU-CCTF), and the RIE2025 Industry Alignment Fund - Industry Collaboration Projects (IAF-ICP) (Award I2301E0026), administered by A*STAR. {\it (Corresponding author: Xiangwang Hou.)}}
\thanks{G. Xie and J. Xu are with the Tsinghua Shenzhen International Graduate School, Tsinghua University, Shenzhen, 518055, China. E-mail: $\{$xgw24, xjzh23$\}$@mails.tsinghua.edu.cn.}
\thanks{Z. Zhang is with the School of Engineering, WestLake University, Zhejiang, 310030, China, Email: stevezhangz@163.com.}
\thanks{X. Hou and Y. Ren are with the Department of Electronic Engineering, Tsinghua University, Beijing, 100084, China. E-mail: $\{$xiangwanghou@163.com; reny@tsinghua.edu.cn$\}$}
\thanks{D. Ma is with the Ocean College, Zhejiang University, Zhoushan, 316000, China. E-mail: mdf2004@zju.edu.cn.} 
\thanks{S. Zhang is with the Department of Data Science, New Jersey Institute of Technology, State of New Jersey, 07450, USA. E-mail: sz457@njit.edu.}
\thanks{D. Niyato is with the College of Computing and Data Science, Nanyang Technological University, Singapore. E-mail: dniyato@ntu.edu.sg.}
\thanks{$^\dagger$ These authors contributed equally to this work.}

}

\markboth{IEEE TRANSACTIONS ON MOBILE COMPUTING}%
{Shell \MakeLowercase{\textit{et al.}}: A Sample Article Using IEEEtran.cls for IEEE Journals}


\maketitle

\begin{abstract}  
  It is significant to employ multiple autonomous underwater vehicles (AUVs) to execute the underwater target tracking task collaboratively. However, it's pretty challenging to meet various prerequisites utilizing traditional control methods. Therefore, we propose an effective two-stage learning from demonstrations training framework, FISHER, to highlight the adaptability of reinforcement learning (RL) methods in the multi-AUV underwater target tracking task, while addressing its limitations such as extensive requirements for environmental interactions and the challenges in designing reward functions. The first stage utilizes imitation learning (IL) to realize policy improvement and generate offline datasets. To be specific, we introduce multi-agent discriminator-actor-critic based on improvements of the generative adversarial IL algorithm and multi-agent IL optimization objective derived from the Nash equilibrium condition. Then in the second stage, we develop multi-agent independent generalized decision transformer, which analyzes the latent representation to match the future states of high-quality samples rather than reward function, attaining further enhanced policies capable of handling various scenarios. Besides, we propose a simulation to simulation demonstration generation procedure to facilitate the generation of expert demonstrations in underwater environments, which capitalizes on traditional control methods and can easily accomplish the domain transfer to obtain demonstrations. Extensive simulation experiments from multiple scenarios showcase that FISHER possesses strong stability, multi-task performance and capability of generalization.
\end{abstract}

% 首字母不需要大写

\section{Introduction}
Autonomous underwater vehicles (AUVs) \cite{29} have broad application prospects in underwater rescue, constructing seamless communication networks, and target tracking. However, the maneuverability and detection range of a single AUV are limited, so multi-AUV swarms based on underwater acoustic links \cite{55} are more widely used in target tracking tasks. Compared with a single AUV, multiple AUVs can adopt a more flexible formation strategy, increase the detection range by information sharing, and overcome tracking failures caused by single-point failures or tracking errors \cite{54}. However, the adoption of multiple AUVs also brings challenges. Specifically, multiple AUVs need to control the distance between each other, move forward consistently, or change formation according to scenarios, while avoiding obstacle danger zones \cite{34}, and more if necessary. Therefore, traditional approaches, such as methods based on Lyapunov vector fields, artificial potential field (APF), and model predictive control (MPC), typically have to make lots of mathematical simplification, making them lack generality and practicality.

Due to its strong ability to feature expression and meet demands, reinforcement learning (RL) provides an efficient solution to tackle these requisites and achieve effective tracking. For example, Yang \MakeLowercase{\textit{et al.}}\cite{17} took the original data of sensors as state and directly output control signals such as propeller thrust, which effectively overcomes the complex influence of the underwater environment. Besides, Wang \MakeLowercase{\textit{et al.}}\cite{18} took lots of demands into consideration, such as energy costs and information sharing. These researches have demonstrated the significant utility in target tracking problems. However, there are some challenges when applying RL. To be specific, the performance of agents strongly relies on the design of the reward function. A well-designed reward function must have tight monotonic correlations with the optimization objectives, which is hard to be satisfied as the number of objectives or agents increases \cite{56,57}. Otherwise, undesired outcomes, such as sub-optimal policies and reward hacking, may be produced\cite{3}. This is especially detrimental to multi-AUV target tracking with complex requirements. Besides, abundant interactions with the environment are required, which leads to high costs of time and computing resources, or even not feasible because of the high risk\cite{4,50}.

The booming development of learning from demonstrations (LfD) in recent years, especially in imitation learning (IL) and offline reinforcement learning (ORL),  provides feasible solutions to tackle these challenges\cite{43,10557650}.  IL requires the policy to learn to perform a task from limited demonstrations. The mainstream IL methods currently comply with the perspective of generative adversarial IL (GAIL)\cite{5}, which introduces a discriminator to align the policy with demonstrations. However, it confronts issues such as low sample efficiency and poor generalization performance. Furthermore, ORL is a widely-known variation of RL that aims to obtain optimal policy given a limited dataset with possibly sub-optimal trajectories without additional interactions\cite{7}. ORL methods possess strong stability and comprehensive performance, but with drawbacks such as demands on the scale of the dataset, and deadly triangle (bootstrap, off-policy and approximation) in estimating the Q-function\cite{8}. To make matters worse, it does not address the inherent dependency of RL on designing reward functions. 

In this paper, we construct a sample-efficient LfD training framework, FISHER, which is dedicated in the multi-AUV target tracking task and complex underwater environment. FISHER exploits the predominant potential of RL in multi-objective optimization, while integrating IL and ORL to avoid reward function design. It avoids the careful trade-offs required for multiple AUVs and complex user requirements to obtain stable and optimal policies. Our main contributions lie in the following:
\begin{itemize}
\item{To the best of our knowledge, we first employ an expert data-driven approach to execute underwater multi-AUV target tracking tasks effectively. We propose an end-to-end, easy-to-deploy framework with a simulation-to-simulation (sim2sim)-based procedure to generate expert demonstrations easily. It consists of IL as the first stage, and ORL as the second stage. The former enables efficient policy improvement with few-shot demonstrations, while the latter further enhances the policies' generalization and multi-task capabilities.}
\item{In the first stage of FISHER, a sample efficient IL algorithm, discriminator-actor-critic (DAC), is introduced, which leverages the replay buffer, off-policy RL algorithm, and improvements for training discriminator to tackle challenges faced by GAIL-based algorithms. Then, we derive the optimization objective of the multi-agent IL algorithm, based on Nash equilibrium and solving the dual optimization problem, thus expanding DAC to multi-agent DAC (MADAC).}
\item{In the second stage of FISHER, a reward function-irrelevant ORL algorithm, multi-agent independent generalized decision transformer (MAIGDT) is introduced. By learning features of state transition from the future, with the help of the hindsight information matcher (HIM), MAIGDT can replicate the demonstrations without prior knowledge.  Comparative experiments and performance evaluations of target tracking tasks show that MAIGDT significantly outperforms RL and ORL methods. Furthermore, our reward-independent workflow offers significant advantages in dense-obstacle scenarios and can be extended to a greater number of AUVs that are challenging for previous MARL approaches.}
\end{itemize}

% 注意related work需要用过去时态，剩下的用现在时态
\section{Related Work}
\subsection{Multi-agent Target Tracking}

% NOTE 段间记得空一行
Numerous methods have been proposed for target tracking tasks. Muslimov \MakeLowercase{\textit{et al.}} designed a decentralized Lyapunov vector field for the target following \cite{32}. Shen \MakeLowercase{\textit{et al.}} deployed a nonconvex programming algorithm into a federated learning framework to optimize performance metrics under communication and latency constraints jointly \cite{31}. In \cite{30}, a neural network-based predictor was introduced to improve the stability of APF and guarantee the Lyapunov stability of obstacle avoidance and connectivity-preserving in target tracking tasks. \cite{28} and \cite{29} constructed a grid diagram-based topologically organized biological neurodynamics model to characterize dynamic environments, which guides AUVs to avoid obstacles and search the target.

Unfortunately, most of these works only apply to ideal and specifically designed environments, thus lacking general applicability due to the highly dynamic underwater environment and complex demands of the tasks \cite{42,10472660}.

\subsection{Multi-agent RL-assisted Target Tracking}

In order to tackle more complex scenarios and user requirements, multi-agent RL (MARL) is increasingly applied to the target tracking task. Wei \MakeLowercase{\textit{et al.}} brought adversarial behaviors between followers and the target into the differential game framework and used MATD3 to optimize the policies, where the system can asymptotically approach Nash Equilibrium \cite{33}. Xia \MakeLowercase{\textit{et al.}} took spatial information entropy into account and utilized the MASAC algorithm, notably increasing the tracking success rate \cite{34}. Yue \MakeLowercase{\textit{et al.}} factorized the centralized critic network of MASAC to reduce the variance in policy updates and learn efficient credit assignments \cite{35}. In \cite{36}, coronal bidirectionally coordinated prediction networks were deployed to MADDPG, aiming to imitate human thinking. 

However, multiple AUVs bring out more trade-offs, making the reward function difficult to design, so most of the existing methods are only for environments with no obstacles and little disturbance. Few existing papers overcome this challenge. For example, Wang \MakeLowercase{\textit{et al.}} implement the classification mechanism-based formation strategies to reduce the coordination cost under multiple AUVs \cite{52}, while Chen \MakeLowercase{\textit{et al.}} deeply integrate the dynamic characteristics of unmanned aerial vehicles (UAVs) into a virtual environment generator combined with the target prediction network to better cope with unseen scenes \cite{53}. Nevertheless, these methods still rely on extensive prior knowledge, so once some conditions change, the reward function and other prior knowledge-based designs must be modified concurrently, severely hindering their application.

\subsection{RL-assisted Task With Demonstrations}

In previous works, expert demonstrations usually enhance the policy training process. In \cite{39} and \cite{40}, classical controller-generated trajectories were mixed into the replay buffer to stabilize the early training stage.  Stev{\v{s}}i{\'c} \MakeLowercase{\textit{et al.}} utilized the MPC controller as an expert to pre-train the policy \cite{38}.

The most similar work to our own is TSDRL-EE from Wang \MakeLowercase{\textit{et al.}} \cite{37}, which adopted TD3 algorithm with behavior cloning (TD3+BC) as first-stage imitation pre-training and self-evolving TD3 that screens excellent experience from replay buffer as the second stage. However, offline RL algorithms like TD3+BC have intense demands on the dataset scale, otherwise poor outcomes may be produced \cite{41}. Besides, the expert-assisted RL training paradigm does not resolve the dependency on the reward function.

Different from these works, our proposed framework is reward function irrelevant, based on few-shot demonstrations. Besides, our sim2sim procedure does not require that the environment of demonstration and policy interaction be the same, notably facilitating expert trajectory generation.

\section{System Model}

\begin{figure}[!t]
  \centering
  \includegraphics[width=1.0\linewidth]{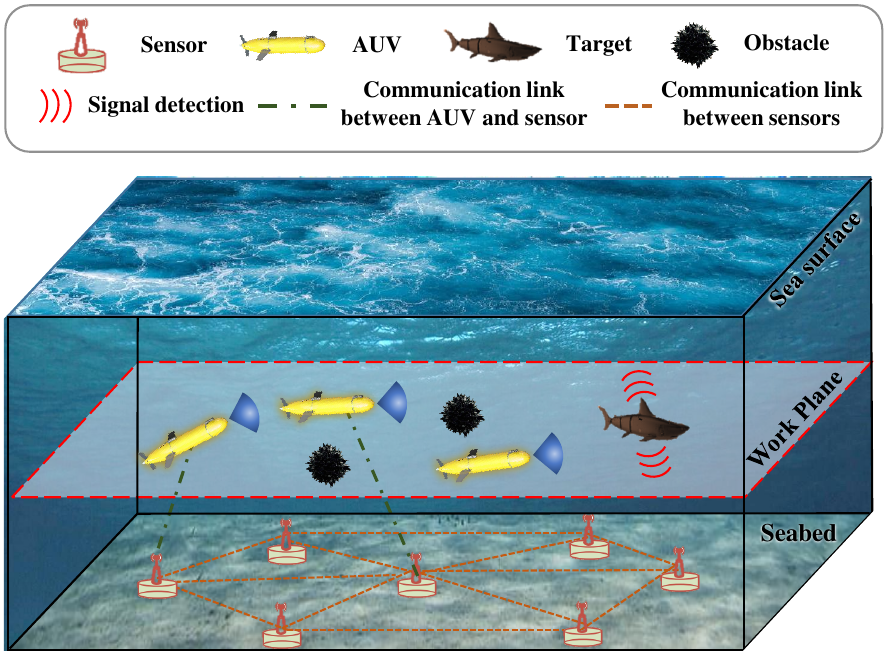}
  \caption{Illustration of the multi-AUV underwater target tracking task.}
  \label{fig_1}
  \end{figure}

In this section, we describe the AUV dynamic model, underwater detection model, action consistency, and Markov decision process (MDP). We consider the system model of the multi-AUV underwater target tracking task as shown in Fig. \ref{fig_1}, $N(N>1)$ AUVs are responsible for tracking a target and moving on the same plane at $d$ meters below the surface. The positions of the target and AUVs are denoted as $\boldsymbol{p}_T=\left[x_T\left(t\right), y_T\left(t\right)\right]$ and $\boldsymbol{p}_i=\left[x_i\left(t\right), y_i\left(t\right)\right]$, where $i\in\{1,2,\cdots,N\}$. There are $M$ obstacles $\{o_1,...,o_M\}$ in the target tracking area, and AUVs should cooperatively track the target while guaranteeing these obstacles are away from the safe radius of AUVs. 

% 此处插入图片

\subsection{AUV Dynamic Model}\label{3A}
Given that each AUV tracks target in the horizontal plane, we can express the dynamic model using a three-degree-of freedom underdrive model, with a body-referenced frame $\boldsymbol{v}_i=\left[v_{i,x}\left(t\right),v_{i,y}\left(t\right),w_i\right]$ and a world-referenced frame $\boldsymbol{\eta}_i=\left[x_i\left(t\right),y_i\left(t\right),\theta_i\right]$, where $v_{i,x}\left(t\right),v_{i,y}\left(t\right),w_i$ and $\theta_i$ represent the surge velocity, sway velocity, angular velocity and yaw angle of the AUV $i$, respectively. Here we adopt the simplified Fossen's dynamic model\cite{48}, and the kinetic equation of AUV $i$ can be expressed as
% 注意：IEEE全部的公式必须以\begin{equation}为起始和终止
\begin{equation}
\boldsymbol{M}_i \dot{\boldsymbol{v}}_i+\boldsymbol{C}_i\left(\boldsymbol{v}_i\right) \boldsymbol{v}_i+\boldsymbol{D}_i\left(\boldsymbol{v}_i\right) \boldsymbol{v}_i+\boldsymbol{G}_i \boldsymbol{\eta}_i=\boldsymbol{\tau}_i,
\end{equation}
where $\boldsymbol{M}_i$ represents the inertia matrix including the additional mass of AUV $i$, while $\boldsymbol{C}_i$ denotes the Coriros centripetal force matrix of AUV $i$, and $\boldsymbol{D}_i$ stands for the damping matrix caused by viscous hydrodynamic. Besides, $\boldsymbol{G}_i$ is the composite matrix of gravity and buoyancy, and $\boldsymbol{\tau}_i$ denotes the control input of AUV $i$. We assume that maintaining depth using a controller is relatively easy for AUVs, namely AUVs can maintain motion in a plane and ensure force balance in the depth direction. The relation between $\boldsymbol{v}_i$ and $\boldsymbol{\eta}_i$ can be expressed by the kinematic equation
\begin{equation}
\dot{\boldsymbol{\eta}}_i=\boldsymbol{J}\left(\boldsymbol{\eta}_i\right) \boldsymbol{v}_i,
\end{equation}
where the transformation matrix $\boldsymbol{J}$ is given by
\begin{equation}
  \boldsymbol{J}\left(\boldsymbol{\eta}_i\right)=\left[\begin{array}{ccc}
    \cos \theta_i & -\sin \theta_i & 0 \\
    \sin \theta_i & \cos \theta_i & 0 \\
    0 & 0 & 1
    \end{array}\right].
\end{equation}

To be applied in simulation, the kinematic and kinetic equations above are discretized over time, and we can obtain
\begin{equation}
  \boldsymbol{\eta}_{t+1}=\boldsymbol{\eta}_t+\Delta T \cdot \boldsymbol{J}\left(\boldsymbol{\eta}_t\right) \boldsymbol{v}_t,
\end{equation}
\begin{equation}
  \boldsymbol{v}_{t+1}=\boldsymbol{v}_t+\Delta T \cdot \boldsymbol{M}^{-1} F\left(\boldsymbol{\eta}_t, \boldsymbol{v}_t\right),
\end{equation}
where $F\left(\boldsymbol{\eta}_t,\boldsymbol{v}_t\right)=\boldsymbol{\tau}_t-\boldsymbol{C}\left(\boldsymbol{v}_t\right)\boldsymbol{v}_t-\boldsymbol{D}\left(\boldsymbol{v}_t\right)\boldsymbol{v}_t-\boldsymbol{G}\boldsymbol{\eta}_t$, and ${\Delta T}$ is the time interval.

\subsection{Underwater Detection Model} % 副标题大写

AUVs use sonar to detect the target and obstacles in the environment. The attenuation of underwater acoustic propagation can be specified by the active sonar equation
\begin{equation}
\label{feess}
  EM=SL-2TL+TS-(NL-DI)-DT.
\end{equation}

All parameters in Eq. \eqref{feess} are in dB, where $SL$, $TL$, $TS$, $NL$, and $DI$ represent the emission sound strength, transmission loss, target strength related to the target reflection area, environmental noise level and directionality index, respectively. $DT$ and $EM$ are the sonar's detection threshold and echo margin, respectively. % 可能要改

For AUV-to-AUV communication, it is only necessary for an AUV to receive the signal from another one, which can be modeled using the passive sonar equation
\begin{equation}
  \label{feepp}
  EM=SL-TL-NL+DI-DT.
\end{equation}

The transmission loss $TL$ is related to the AUV-target distance $d$ and center acoustic frequency $f$, i.e.
\begin{equation}
  TL=20 \lg (d)+d \times \alpha(f) \times 10^{-3},
\end{equation}
\begin{equation}
  \alpha(f)=0.11 \frac{f^2}{1+f^2}+44 \frac{f^2}{4100+f^2}+2.75 \times 10^{-4} f^2+0.003,
\end{equation}
where $\alpha\left(f\right)$ is the empirical formula for the attenuation of sound waves in water. Since $EM$ and $d$ show a monotonically decreasing relationship, the maximum detection radius $r_c$ of an AUV is
\begin{equation}
r_c={\mathop{\rm{argmax}}\limits_{d}}\{EM(d)\geq0\}.
\end{equation}

Given that the transmission loss of the passive sonar equation is only from the one-way propagation loss between AUVs rather than the two-way in the active equation, we consider that the communication range between AUVs significantly exceeds the tracking distance from the AUV to the target, namely, that AUVs' communication will be available.

% \section{Problem Formulation}
% In this section, we first model the interaction between AUVs and the target as a Markov decision Process(MDP). Then, the state space, action space and reward function are designed in detail.
\subsection{Action Consistency}

A high swarm consistency means that AUVs can track the target jointly. However, as the number of AUVs increases, it is hard to express the consistency directly from mutual distances between AUVs. Here, we use topology connectivity among AUVs to define the consistency. Similar to Eq. \eqref{feepp}, we define the signal-to-noise ratio (SNR) between AUV $i$ and AUV $j$ in underwater communication
\begin{equation}
  a_{ij} = SL - TL - NL + DI.
\end{equation}

Then we formulate the AUV swarm as a graph and utilize the Laplace matrix $L \in \mathbb{R}^{N\times N}$ to describe the consistency of the AUV swarm. The element in the $i$-th row and $j$-th column is defined as follows:
\begin{equation}
  l_{ij} = \begin{cases}
    -a_{ij},  & i \neq j, a_{ij} \geq DT, \\
    \sum^{k=N}_{k=1,k\neq i}a_{ik}, & i=j, a_{ik} \geq DT, \\
    0,   & i \neq j, a_{ij} < DT.
  \end{cases}
\end{equation}

Thereby, the algebraic connectivity is the second smallest eigenvalue $\lambda$ of matrix $L$. Larger $\lambda$ predicates stronger consistency of the swarm.

\subsection{Markov Decision Process}

We model the interaction between AUVs and the target as an MDP, which assumes that the actions of AUVs only depend on the current state of the environment. This MDP can be expressed as a quintuple
\begin{equation}
  \Omega = (\boldsymbol{S},\boldsymbol{A},\boldsymbol{P},\boldsymbol{R},\gamma),
\end{equation}
where $\boldsymbol{S}$ represents the state space, and $\boldsymbol{A}$ denotes the action space of AUVs. Besides, $\boldsymbol{P}$ stands for the state transition probability function, and $\boldsymbol{R}$ represents the reward function, and $\gamma$ is the discount factor. To be specific, the details of each element in the tuple can be listed as follows:

\textit{1) State space $\boldsymbol{S}$:} The $i$-th AUV's state $\boldsymbol{s}_i\left(t\right)\in\mathbb{R}^{4N+2N_o}$ in the state space $\boldsymbol{S}_i$ is the concatenation of these parts

  \begin{itemize}
\item{\textbf{Target's position and velocity:}} This part is signified by formula below:
\begin{equation}
  \boldsymbol{s}_{i_1}(t)=\left\{x_{i, t}(t), y_{i, t}(t), v_{x_{i, t}}(t), v_{y_{i, t}}(t) \right\},
\end{equation}
where $x_{i, t}(t)$ and $y_{i, t}(t)$ are the relative positions of the target to the AUV, while $v_{x_{i, t}}(t)$ and $v_{y_{i, t}}(t)$ are the relative velocities. These values are defined in the coordinate system of the polar axis in which the direction of the $i$-th AUV is facing, namely $x_{i, t}(t) = d_i(t)\cos(\theta_{i,t}(t))$, the same applies hereinafter. 
\item{\textbf{Other AUVs' position and velocity:}} This part is defined similarly. It is worth noting that this part includes the position and velocity information of all other AUVs
\begin{equation}
  \begin{aligned}
  & \boldsymbol{s}_{i_2}(t)=\left\{x_{i,j}(t),y_{i,j}(t),v_{x_{i,j}}(t),v_{y_{i,j}}(t) \right. \\
  & \left.\qquad\qquad\qquad\qquad\quad | \ j \in \{1,...,N\} \setminus \{i\}  \right\}.
  \end{aligned}
\end{equation}
\item{\textbf{Obstacles' position:}} It is assumed that the AUV swarm can detect at most $N_o$ obstacles, and this part is defined as
\begin{equation}
  \begin{aligned}
    & \boldsymbol{s}_{i_3}(t)=\left\{EM_j\cos(\theta_{io_j}(t)), EM_j\sin(\theta_{io_j}(t))\right. \\
    & \left.\qquad\qquad\qquad\qquad\qquad | \ j \in \{1,...,N_o\} \setminus \{i\}   \right\},
\end{aligned}
\end{equation}
where $EM_j$ is the echo margin of obstacle $o_j$, while the angle between $o_j$'s position relative to the AUV and AUV's orientation is defined as $\theta_{io_j}$. When less than $N_o$ obstacles are detected, the according $EM$ is set to 0dB.
\end{itemize}

\textit{2) Action Space $\boldsymbol{A}$:} 
The action space for each AUV is defined through its motion constraints as established in Section \uppercase\expandafter{\romannumeral3}-A. For the $i$-th AUV in the MDP framework, the action space and corrsponding action at time $t$ are defined as:
    \begin{equation}
        \boldsymbol{A}_i = \big[0,\, v_{\max}\big] \times \big[-\omega_{\max},\, \omega_{\max}\big],
    \end{equation}
    \begin{equation}
        \boldsymbol{a}_i(t) = \big[v^{\text{des}}_i(t),\, \omega^{\text{des}}_i(t)\big],
    \end{equation}
where $v_{\max}$ and $\omega_{\max}$ represent the maximum linear and angular velocities, respectively. $v^{\text{des}}_i(t) \in [0, v_{\max}]$ and $\omega^{\text{des}}_i(t) \in [-\omega_{\max}, \omega_{\max}]$ denote the desired velocity and desired angular velocity, respectively. These are computed from the DRL model and differ from the actual velocities governed by the dynamic model. Then, the next state $\boldsymbol{s}_i(t+1)$ is determined by the joint actions of all $N$ AUVs through the state transition probability: $\boldsymbol{s}_{i}(t) \times \boldsymbol{a}_{1}(t) \times \cdots \times \boldsymbol{a}_{N}(t) \longmapsto 
\boldsymbol{P}_i\big(\boldsymbol{s}_i(t+1) \,|\, \boldsymbol{s}_i(t),\, \boldsymbol{a}_1(t), \ldots, \boldsymbol{a}_N(t)\big),$
which models the stochastic interactions between the AUVs and their environment.

\textit{3) Reward function $\boldsymbol{R}$:} As mentioned before, the reward function should be highly correlated with our requirements of the target tracking task. Still, it is not easy to achieve this, especially for complex scenarios. Therefore, we only utilize the reward function as an indicator to measure performance in simple scenarios, utilize a typical RL algorithm for training, and compare it with FISHER in the later experimental section.  For the FISHER framework, the rewards of the MDP are derived from latent variables, which improves the policy to approximate the expert demonstrations. The details will be discussed in Section \uppercase\expandafter{\romannumeral4}.

Here, the reward function $r_i(t) \in \boldsymbol{R}$ of $i$-th AUV consists of optimization objectives that correspond to our demands, which are listed as follows:
\begin{equation}
  \label{r1}
  r_{t i_i}(t)= \begin{cases}d_i(t)-d_{\min }^t(t), & d_i(t)>d_{\min }^t, \\ 0, & d_i(t)<d_{\min }^t,\end{cases}
\end{equation}
\begin{equation}
  \label{r2}
  \begin{aligned}
  & r_{o_i}(t)\!=\!\!\!\sum_{\substack{j=1, j \neq i}}^N\!\!\!\left(d_{\text{safe}}\!-\! d_{i j}(t)\right)\!+\!\!\!\sum_{k=1, k \neq i}^M\!\!\!\left(d_{\text{safe}}\!-\! d_{i, o_k}(t)\right), \\
  & \qquad\qquad\qquad \text { for all } d_{i j}(t)\!<\! d_{\text{safe}}, \ d_{i, o_k}(t)\!<\! d_{\text{safe}},
  \end{aligned}
\end{equation}
\begin{equation}
  \label{r3}
  r_{l_i}(t)= \begin{cases}
  \lambda_{0} - \lambda_{\max}, & \lambda(t) \geq \lambda_{\max}, \\
  \lambda_{0} - \lambda(t), & \lambda(t) <  \lambda_{\max}.
  \end{cases}
\end{equation}

The meaning of each term in Eq. \eqref{r1} $\sim$ \eqref{r3} are elaborated as follows:

\textit{1) \textbf{Target tracking reward:}} $r_{ti_i}$ is determined by the distance between $i$-th AUV and the target, aiming at encouraging a single AUV to track the target independently, where $d^t_{\min}$ is the optimal distance from the target. We also introduce a term $r_{t c}(t)=\max _i\left\{r_{t i_i}(t)\right\}$ to represent overall tracking performance. 

\textit{2) \textbf{Collision avoidance penalty:}} $r_{o_i}$ is used to avoid collision with all other AUVs and obstacles. This penalty is from all AUVs and obstacles that are closer than the safe distance $d_{\text{safe}}$ from the current AUV, and all the penalties will be summed up.  

\textit{3) \textbf{Swarm consistency reward:}} This reward is directly from the algebraic connection $\lambda$. As the value of $\lambda$ is usually much larger than 0, we offset $\lambda$ with a constant $\lambda_0$. Excessive $\lambda$ is truncated to avoid collisions.

To compare with the proposed FISHER framework, while accent the limitations of designing the reward function, we set two weight factors, $a$ and $b$ for $r_{ti_i}$ and $r_{tc}$, to adjust the positivity of AUVs tracking target. Here, we propose three settings: 
\textbf{Cooperative:} $a=1, b=0$; \textbf{Mixed}: $a=0.5, b=0.5$; \textbf{Split}: $a=0,b=1$. The cooperative setting only requires that at least one AUV approach the target, while the split setting encourages each AUV to maintain proximity unilaterally for the consideration of robustness. Simulation results will be detailed in Section \uppercase\expandafter{\romannumeral5}.

Then, the reward function $r_i$ of $i$-th AUV can be given by
\begin{equation}
  r_i(t) = w_1(ar_{tc}(t)+br_{ti_i}(t)) + w_2r_{o_i}(t) + w_3r_{l_i}(t),
\end{equation}
where $W=[aw_1,bw_1,w_2,w_3]$ is the weight vector.

\section{Methodology} 
In this section, we detail the expert demonstration-based training framework FISHER for target tracking. First, we introduce our sim2sim method, which aims to simplify the generation of expert trajectories. Then, we demonstrate the two stages of FISHER in order, followed by the 
detailed description of the overall architecture of the proposed FISHER framework.

\begin{figure}[!t]
  \centering
  \includegraphics[width=0.948\linewidth]{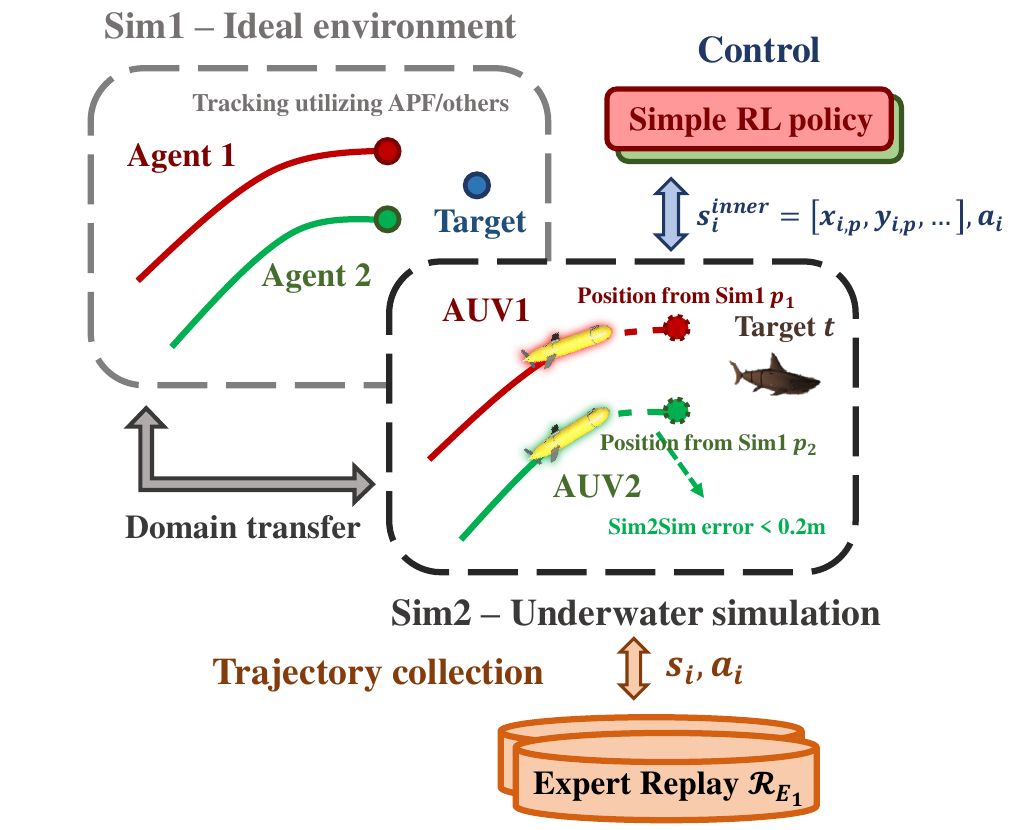}
  \caption{A simple illustration of sim2sim procedure.}
  \label{figgg}
  \end{figure}

\subsection{Sim2sim Expert Demonstration Generation}

It is intractable to directly generate demonstrations in underwater (simulation) environment due to the significant complexity, while RL methods suffer from the drawbacks of designing reward functions. Therefore, we propose a sim2sim procedure to simplify this process. The overall procedure is illustrated in Fig. 2.

To be specifical, our sim2sim method consists of these parts: 

\textit{1) \textbf{Demonstration in simplified environment:}} We first simplify the environment by ignoring underwater and other environmental effects and modeling AUVs and the target as particles (denoted as \textbf{sim1}). This abstraction permits direct application of classical tracking approaches, such as by APF \cite{9}, to directly control AUVs' position without incorporating actuator and environmental constraints. This approach enables traditional tracking methodologies to access sufficient information (e.g., precise positions of the target and obstacles), which may not be fully attainable in practical deployments.  

\textit{2) \textbf{Sim2sim:}} To do this, we train a one-by-one tracking RL policy, whose sole function is to allow AUV to reach the specific waypoint with a specific velocity in the simulated environment with disturbance (denoted as \textbf{sim2}). 

Specifically, the state space of the internal RL policy consists of the AUV's position and orientation, as well as the target's position and velocity, while its action space is identified as described in Section \uppercase\expandafter{\romannumeral3}. The reward function for the internal RL policy consists of the negative Euclidean distance to the target point and the negative mean square error (MSE) of target velocity.  Due to the simplicity of the training objective, the tracking error can quickly converge to $<0.2\text{m}$.   

\textit{3) \textbf{Expert buffer collection:}}
The RL policy derived from 2) is deployed to each AUV to execute the target tracking task in sim2 under the guidance of demonstrations generated in sim1. The generated execution trajectories are recorded, with their state space and action space aligning with those described in Section \uppercase\expandafter{\romannumeral3}, thereby ensuring direct applicability in subsequent workflows. Besides, some measures are taken to enhance the diversity of the dataset, which will be detailed in the Section V.

\subsection{Improved Imitation Learning}

GAIL is an important IL method that guides the policy in approximating the demonstrations by training a discriminator to distinguish between expert and policy-generated trajectories. To achieve this, GAIL employs the maximum entropy inverse RL (IRL) framework to obtain the reward function, and then the expert policy can be derived via RL procedure\footnote[1]{\cite{5} uses the cost function $c:\boldsymbol{S}\times\boldsymbol{A}
\to\mathbb{R} $ to notate these optimization objectives. The definition of the cost function is the opposite of the reward function.}
\begin{equation}
  \label{gail1}
\text{RL}(r) = {\mathop{\rm{max}}\limits_{\pi \in \Pi}}\; H(\pi) + \mathbb{E}_\pi[r(s, a)], 
\end{equation}
\begin{equation}
  \label{gail2}
  \begin{aligned}
     \text{IRL}_{\psi}(\pi_E) & = \mathop{\rm{argmax}}\limits_{r \in \mathbb{R}^{\boldsymbol{S} \times \boldsymbol{A}}}-\psi(r) + \mathbb{E}_{\pi_E}[r(s, a)] \\
    & - \left(\max_{\pi \in \Pi} H(\pi) + \mathbb{E}_{\pi}[r(s, a)]\right),
  \end{aligned}
\end{equation}
where $H(\pi)=\mathbb{E}_{s_t,a_t\sim \pi }\left[-\sum_{t=0}^\infty\gamma^t\log \pi(a_t| s_t)\right]$ denotes the $\gamma$-discounted casual entropy, $\pi_E$ is the expert policy, and $\psi$ is a reward function regulizer [8]. According to Ho. \MakeLowercase{\textit{et al.}}\cite{5}, we can obtain the dual optimum
\begin{equation}
  \label{gail3}
  \text{RL} \circ \text{IRL}_\psi(\pi_E) = \mathop{\rm{argmin}}\limits_{\pi \in \Pi} -H(\pi) + \psi^\star(\rho_\pi - \rho_{\pi_E}),
\end{equation}
where $\rho_\pi(s,a)=\pi(a|s)\sum^{\infty}_{t=0}\gamma^t \boldsymbol{P}(s_t=s|\pi)$ stands for policy's occupancy measure. GAIL utilizes a well-designed regulizer $\psi_{GA}$, and final objective can be expressed as
\begin{equation}
  \label{gail4}
  \begin{aligned}
  & \psi_{GA}^{*}(\rho_\pi - \rho_{\pi_E}) =  \mathop{\rm{max}}\limits_{D} \mathbb{E}_{\pi_E}[\log(D(s, a))] \\
  & \qquad\qquad\qquad\quad\;\;+ \mathbb{E}_{\pi}[\log (1 - D(s, a))],
  \end{aligned}
\end{equation}
where $\psi_{GA}^{*}$ is the convex conjugate of $\psi$, and $D:\boldsymbol{S}\times\boldsymbol{A}\to(0,1)$ is the discriminator. Finally, the policy can be improved via on-policy RL algorithms such as TRPO\cite{44} and PPO\cite{6}. However, it is intractable that GAIL demands extensive interactions with the environment. To address this, Kostrikov \MakeLowercase{\textit{et al.}}\cite{10} introduces the replay buffer to store previously generated trajectories. Then, the training objective of the discriminator can be expressed as
\begin{equation}
  \mathcal{L}_D=\mathbb{E}_{\mathcal{R}}\left[\log \left(D(s, a)\right)\right]+\mathbb{E}_{\pi_{E}}\left[\log \left(1-D(s, a)\right)\right],
\end{equation}
where $\mathcal{R}$ denotes the replay buffer of the $i$-th AUV. Then, we can employ off-policy actor-critic algorithms for policy training, such as SAC\cite{26} and TD3\cite{45}, and this training paradigm is named as discriminator actor-critic (DAC).

Besides, we utilize some common improvements for discriminator training, including gradient penalty (GP)\cite{21} and spectral normalization (SN)\cite{22}, which can notably enhance training stability and efficiency. Also, absorbing state $s_a$\cite{20} is introduced to avoid a policy reaching the episode termination positively. Specifically, the absorbing state is entered when an episode is abnormally terminated, and every action transits the state to itself, namely: $r(s_a,\cdot) = 0$, and $\boldsymbol{P}(s_{t+1}=s_a|s_t=s_a,\cdot) = 1$. For consistency and unbiased reward function for absorbing state, internal reward function for training RL algorithm is signified by
\begin{equation}
  r(s,a) \leftarrow \log D(s,a) - \log(1-D(s,a)).
\end{equation}
\subsection{Multi-Agent Discriminator Actor-Critic}

\begin{algorithm}[!htb]  
  \caption{MADAC training \& Offline dataset generation (the first stage of the FISHER framework)}  
  \label{alg:1}  
  \textbf{Initialize:} Replay buffer $\boldsymbol{\mathcal{R}} = [\mathcal{R}_{1},\ldots,\mathcal{R}_{N}]$, expert trajectory\! buffer $\boldsymbol{\mathcal{R}}_E = [\mathcal{R}_{E_1},\ldots,\mathcal{R}_{E_N}]$, \!\!\! discriminator network $D$, policy network $\pi_{\theta_i}$ with corresponding critic network, offline dataset $\mathcal{R}_{o_i}$ of AUV $i$.
  
  \For{each episode ${k}$}{
  
  Reset the training environment.
  
  \For{each environment timestep $t$}{
  Sample action $a_{t_i} \sim \pi_{\theta_i}\left(\cdot \mid s_{t_i}\right).$
  
  Collect the next state $s_{{t+1}_i} \sim \boldsymbol{P}_i(\cdot|\boldsymbol{s},\boldsymbol{\pi}).$
  
  Store transition $\mathcal{R}_i \leftarrow \mathcal{R}_i \cup\left\{\left(s_{t_i}, a_{t_i}, \cdot, s_{t+1_i}\right)\right\}.$
  }
  \For{each IL gradient step}{
  Sample transitions from replay 
  \qquad $\{(\boldsymbol{s}_{t}, \boldsymbol{a}_{t}, \cdot, \cdot)\}_{t=1}^B \!\sim\! \boldsymbol{\mathcal{R}}$, 
  \qquad $\left\{\left(\boldsymbol{s}_{t}^{\prime}, \boldsymbol{a}_{t}^{\prime}, \cdot, \cdot\right)\right\}_{t=1}^B \!\sim\! \boldsymbol{\mathcal{R}}_{E}.$
  
  Calculate loss  
  
  \ \ $\mathcal{L}_{D}\!=\!\sum_{b=1}^B \!\log\! D\left(\boldsymbol{s}_{b}, \boldsymbol{a}_{b}\right)\!-\!\log \left(1\!-\!D\left(\boldsymbol{s}_{b}^{\prime}, \boldsymbol{a}_{b}^{\prime}\right)\right).$
  
  Update $D$ with Adam+GP+SN.
  }
  \For{each RL gradient step}{
  Sample $\left\{\left(s_{t_i}, a_{t_i}, \cdot, s_{t+1_i}\right)\right\}_{t=1}^B \sim \mathcal{R}_i.$
  
  \For{$b=1, \ldots, B$}{
  $r_i \!\!\leftarrow\!\! \log\! D\left(s_{b_i}, a_{b_i}\right)\!-\!\log \left(1\!-\!D\left(s_{b_i}, a_{b_i}\right)\right).$
  
  $\left(s_{b_i}, a_{b_i}, \cdot,{s}_{b+1_i}\right) \leftarrow\left({s}_{b_i}, {a}_{b_i}, r_i, {s}_{b+1_i}\right).$
  }
  Update $\pi_{\theta_i}$ with SAC\cite{26}.
  }
  }
  Collect trajectories $\tau_i$ using optimal policy $\pi^{*}_{\theta_i}$.
  
  Store transition $\mathcal{R}_{o_i} \leftarrow \mathcal{R}_{o_i} \cup \tau_i$. 
  \end{algorithm}

We turn our attention to extending DAC to multi-AUV scenarios. The optimum policies for MARL can be derived from a Nash equilibrium, namely, an agent cannot improve its own policy to achieve higher rewards if other agents keep their policies fixed. It can be expressed as a constrained optimization problem\cite{46}
\begin{equation}
  \begin{aligned}
    \label{mgail}
  & \text{MARL}(\boldsymbol{R}) = \mathop{\rm{argmin}}\limits_{\boldsymbol{\pi}\in\Pi,\boldsymbol{v}}f_{\boldsymbol{r}}(\boldsymbol{\pi},\boldsymbol{v})-H(\boldsymbol{R}), \\
  & \; s.t. \; v_i(\boldsymbol{s}) \geq q_i(\boldsymbol{s},a_i), \; \forall i \in \{1,\cdots,N\},
\end{aligned}
\end{equation}
where $f_{\boldsymbol{r}}(\boldsymbol{\pi}, \boldsymbol{v}) = \sum_{i=1}^{N} \left(\sum_{\boldsymbol{s} \in \boldsymbol{S}} v_i(\boldsymbol{s}) - \mathbb{E}_{a_i \sim \pi_i(\cdot | \boldsymbol{s})} q_i(\boldsymbol{s}, a_i) \right)$, $\boldsymbol{s} = [s_1,\cdots,s_N]$ is the global state, $\boldsymbol{v} \triangleq [v_1,...,v_N]$ denotes the value functions of policies, while $\boldsymbol{q}$ is the corresponding Q-function. If the Nash equilibrium is satisfied, the objective has a minimal value of zero, which is the only solution to the Nash equilibrium\cite{47}.

% 由于公式(30)是通用两种设定的，所以这个公式并没有写错
Accoring to Song \MakeLowercase{\textit{et al.}}\cite{23}, we can finally obtain the objective of training discriminator(s), similar to Eq. \eqref{gail4}
\begin{equation}
  \label{wtproof}
  \mathop{\rm{max}}\limits_{\boldsymbol{D}} \mathbb{E}_{\boldsymbol{\mathcal{R}}}[\sum^{N}_{i=1}\log D_i(\boldsymbol{s},\boldsymbol{a})]+\mathbb{E}_{{\pi}_E}[\sum^{N}_{i=1}\log (1 - D_i(\boldsymbol{s},\boldsymbol{a}))],
\end{equation}
where the proof of Eq. \eqref{wtproof} is defered to the Appendix.

Next, we discuss the discriminator's training paradigms. Similar to MARL, we can train the discriminator in a centralized or decentralized setting. Specifically, the centralized setting utilizes a single discriminator that takes trajectories concatenated with all AUVs and assigns the same score to each AUV, namely $D=D_1=\cdots=D_N$. In contrast, the decentralized setting equips a discriminator for each AUV, i.e. $D_i(\boldsymbol{s},\boldsymbol{a}) = D_i(s_i,a_i)$. As the training stability is crucial to RL training procedure, we employ the centralized setting, which will exhibit more significant benefits as the number of AUVs $N$ grows. The performance difference of the two settings is given in the subsequent section.

In conclusion, MADAC introduces three core improvements over standard GAIL: (1) an off-policy training architecture integrating SAC and replay buffers to address GAIL’s sample inefficiency, enhanced by gradient penalty and spectral normalization for discriminator stability; (2) a modified reward function with absorbing states to prevent premature termination; and (3) multi-agent innovations including a Nash equilibrium-driven objective and centralized discriminator design to extend GAIL’s single-agent formulation to cooperative settings. These enhancements enable MADAC to achieve faster convergence, stable scalability, and learn effectively from sparse demonstrations.

% 搞一个示意图
\begin{figure}[!t]
  \centering
  \subfloat[DT]{\centering\includegraphics[width=1.50in]{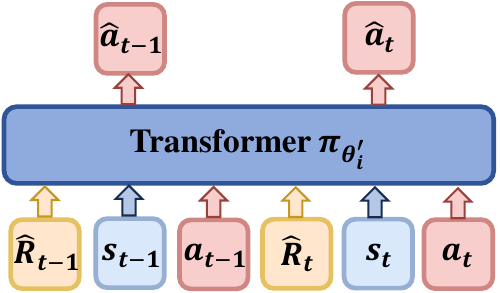}\label{fff1}%
  \label{fig_first_case}}  \hspace{0.10in}
  %\hfil
  \subfloat[MAIGDT]{\centering\includegraphics[width=1.70in]{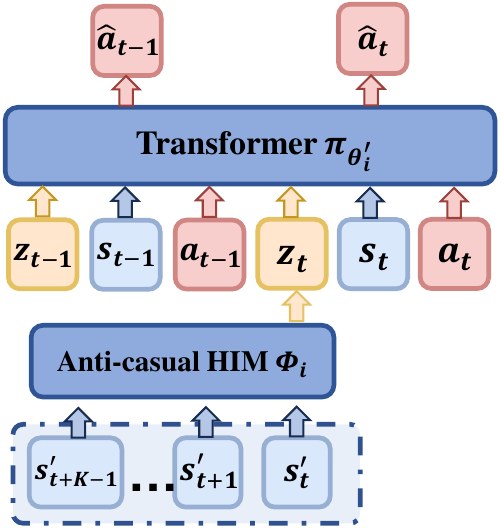}\label{fff2}%
  \label{fig_second_case}}
  \caption{The architecutres of DT\cite{24} and MAIGDT.}
  \label{fig3}
  \end{figure}

% 架构图插入到实验部分之前，不应该与实验部分混同.
% 1.0linewidth表示宽度恰好填满页面。
\begin{figure*}[!t]
  \centering
  \includegraphics[width=1.0\linewidth]{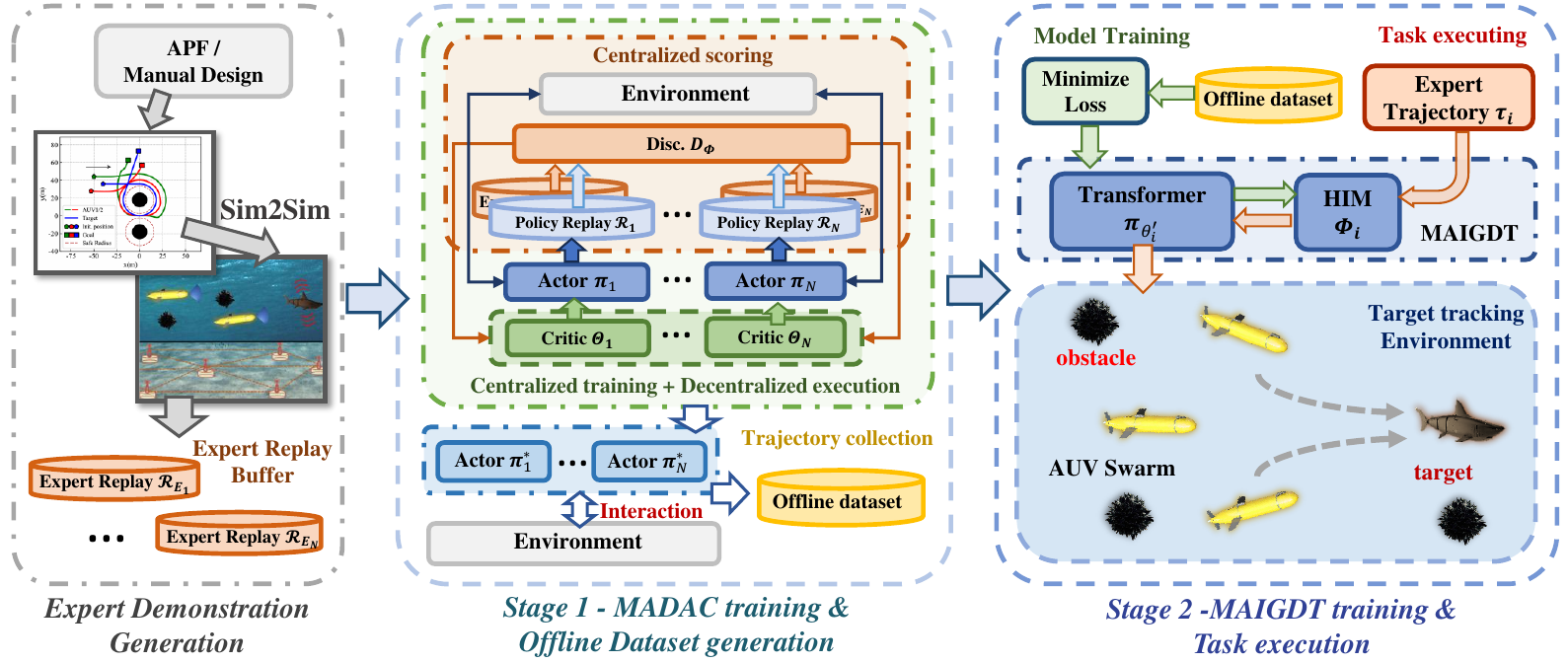}
  \caption{The overall architecture of the FISHER framework in the multi-AUV underwater target tracking task.}
  \label{fig_2}
  \end{figure*}

% 算法亦在此处 \Comment

\subsection{Multi-Agent Independent Generalized Decision Transformer}

Offline RL helps policy improvement without interacting with the environment, among these algorithms, decision transformer (DT)\cite{24} is an important application of generative models in ORL. It abstracts ORL problems into the seq2seq problems, thus eliminating the wrong estimation of Q-function in TD-based ORL algorithms.

To be specific, DT utilizes autoregressive predict-based language models, like GPT-2, to predict action. The GPT-like models utilize a stack of multiple decoders, namely self-attention layers with layer norm (LN) and residual connections. The self-attention layer takes $n$ input tokens as embeddings $\{x_i\}^n_{i=1}$, and outputs embeddings with same dimension $\{z_i\}^n_{i=1}$. Specifically, input tokens are mapped to the key ($k_i$), query ($q_i$) and value ($v_i$) via linear transformations. Output tokens are the weighted average of values, based on the dot product between query and key 
\begin{equation}
  z_i = \sum^{n}_{j=1} \text{softmax}(\{<q_i,k_{j'}>\}^n_{j'=1})_j \cdot v_j.
\end{equation}

Then, DT takes a batch of segments of trajectories with $K$ timesteps, and the modified trajectories as the token sequence. A typical snippet of trajectory from timestep $t$ can be denoted as
\begin{equation}
  \tau_{t_i}^{\prime}=\left(\hat{r}_{i}^{(t)},s_{i}^{(t)}, a_{i}^{(t)}, \ldots, \hat{r}_{i}^{(t+K-1)}, s_{i}^{(t+K-1)}, a_{i}^{(t+K-1)}\right).
\end{equation}

Here, the original DT utilizes the expected return $\hat{r}_{t_i}=\sum^{T}_{t'=t}r_{t'_i}$ of the $i$-th AUV, as an indicator of features of the trajectory. Then, the DT model predicts the next token, based on the input token sequence. Therefore, the prediction head corresponding to the input token $s_i(t)$ is trained to predict $\hat{a}_i(t)$. The training loss of the DT model for each timestep is averaged, namely
\begin{equation}
  \label{DT2}
  {\mathop{\rm{max}}\limits_{\pi_{\theta_i'}}}J'(\theta_i')={\mathop{\rm{min}}\limits_{\pi_{\theta_i'}}}\mathcal{L}_{\text{MSE}}(\theta_i')={\mathop{\rm{min}}\limits_{\pi_{\theta_i'}}}[-\frac{1}{B}\sum^{B}_{j=1}(a_j-\hat{a}_j)^2].
\end{equation}

With the utilization of transformer and autoregressive training, DT is able to match trajectories with high returns based on credits assigned by self-attention layers. However, this training paradigm still relies on the reward function. Besides, as expected return is the only indicator of expected trajectories, multitasking performance can be poor or not feasible. 

% 算法放置处
% 拆分！
\begin{algorithm}[!t]  
  \caption{MAIGDT training \& Task executing (the second stage of the FISHER framework)}  
  \label{alg:1}  
  \textbf{Initialize:} Offline dataset $\mathcal{R}_{o_i}$, DT model parameters $\theta_i^{\prime}$ with ialgorithmts anti-casual transformer $\Phi_i$ of AUV $i$.
  
  Sample ${n}$ batches of sequence with length ${K}$ from the offline dataset $\tau_i$.

  \For{each GDT gradient step}{
  Flip the state of sequences and get ${z}_i$ vectors from anti-casual transformer ${\Phi}_i$.

  Update models of GDT by Adam updating on ${\Phi}_i$ and $\theta_i^{\prime}$ by $L_{\mathrm{MSE}}\left(\theta_i^{\prime}\right)$ of Eq. \eqref{DT2}.
  }

  Get expert demonstration $\tau^{\prime}_{E_i}$ for imitation.

  \While{target tracking task timestep $t$}{
  Get flipped state sequence from timestep ${t}+{K}-1$ to ${t}$ of $\tau^{\prime}_{E_i}$, and get ${z}_{t_i}$ vector from anti-casual transformer ${\Phi}_i$.

  Predict action based on vector $z_i$, state $s_i$ and $a_i$ of previous ${K}$ timesteps.
  }
  \end{algorithm}

Furuta \MakeLowercase{\textit{et al.}}\cite{25} have proposed the generalized decision transformer (GDT) and demonstrated that we can use other information, rather than expected return, to find positive examples with certain contextual parameter values as a hindsight information matcher (HIM). Thus, we can improve DT to match the state transition of selected demonstrations to predict actions. 
In particular, we can use a second transformer $\Phi$, which utilizes the anticasual design, namely takes a reverse-order state sequence as the input. The output of transformer $\Phi$ is the vector $z$ that contains the information of future state transitions. Given that $\Phi$ is differentiable to DT's action-prediction loss, $\Phi$ can learn sufficient features of states by optimizing the Eq. \eqref{DT2}, and DT is proficient in matching any distribution to an arbitrary precision. Namely, we can achieve the objective as:

 \begin{equation}
  \mathop{\text{argmin}}\limits_{\pi} -D(\rho^\pi(s),\rho^*(s)),
\end{equation}
where $D$ is the divergence (e.g., Kullback-Leibler (KL) divergence), $\rho^\pi(s)=\sum^{\infty}_{t=0}\gamma^t\boldsymbol{P}(s_t=s|\pi)$ is the state occupancy measure, and $\rho^*(s)$ is the target state distribution\cite{lee2019efficient}.

Then, when executing target tracking tasks, we can specify an expert trajectory $\tau'_E$ and use $\Phi$ to get features, which guides DT to replicate the one-shot demonstration efficiently. To be intuitive, the architectures of DT and MAIGDT are illustrated in Fig. 3.  

Based on GDT and different from MADAC, we extend GDT to MAIGDT by implementing a decentralized training setting here, thanks to the powerful generalization capabilities of transformer-based models, and minor perturbations do not significantly affect the performance. We further demonstrate the stability of MAIGDT in Section V. Also, the decentralized setting contributes more to the deployment and scalability to policies.

\subsection{The Overall Architecture of The FISHER Framework}

\begin{table} % table for 1/2 width(one column), table* for full width (two columns)
  \caption{Parameters of The Environment and Algorithm.}
  \label{tab:arc-cifar}
  %  	\resizebox{1.0\linewidth}{!}{
  \begin{tabular}{p{0.17\columnwidth}|p{0.45\columnwidth}|p{0.22\columnwidth}}
    \toprule
    & \textbf{Parameters} & \textbf{Values} \\
    \midrule
    \multirow{4}{*}{\makecell[l]{\textbf{Environment} \\ \textbf{Parameters}}}
    & \makecell[l]{Hydroacoustic parameters \\ $SL$, $TS$, $DI$, $DT$, $NL$} & \makecell[l]{$100$dB,$3$dB,$3$dB,\\$20$dB,$30$dB}\\ %[-4pt] 
    & Transmit frequency $f$ & $10\text{kHz}$\\
    & Maximum speed $v_{\max}$   & $2.4\text{m/s}$\\ 
    & Maximum angular speed $w_{\max}$   & $1.0\text{rad/s}$\\
    \midrule
    % \multirow{4}{*}{\textbf{Inference}} & $[\textrm{Fully-connected}+\textrm{Conv}]$ / NVD& $512\times 4 \times 4$\\
    \multirow{10}{*}{\makecell[l]{\textbf{Algorithm} \\ \textbf{Parameters}}}& 
    \makecell[c]{\vspace{-0.5em} \\Reward weight factor  $w_1,w_2,w_3$ \\ \vspace{-0.5em}}  & \makecell[l]{$-0.25$,$-0.4$,\\$-0.2/N$}\\
    & Distance parameters  $d^t_{\text{min}},d_{\text{safe}}$     & $12$m,$8$m\\
    & Consistency parameters $\lambda_{\max},\lambda_0$     &  $52N$,$50N$  \\
    & Hidden layer size           &   $256$     \\
    & batch size                  &   $256$   \\
    & Discount factor $\gamma$    &   $0.99$    \\
    & Learning rate of SAC/MADAC &   $3\times 10^{-4}$       \\
    & Learning rate of MAIGDT  &   $1\times 10^{-4}$        \\
    & MADAC gradient penalty factor &   $1.0$     \\
    & MAIGDT context length $K$       &   $20$      \\
    & MADAC number of demonstration episodes per scenario &  $50$(Generalization experiment), $10$(Others)      \\
    \bottomrule
  \end{tabular}%}
\end{table}

\begin{figure}[!t]
  \centering
  \includegraphics[width=0.98\linewidth]{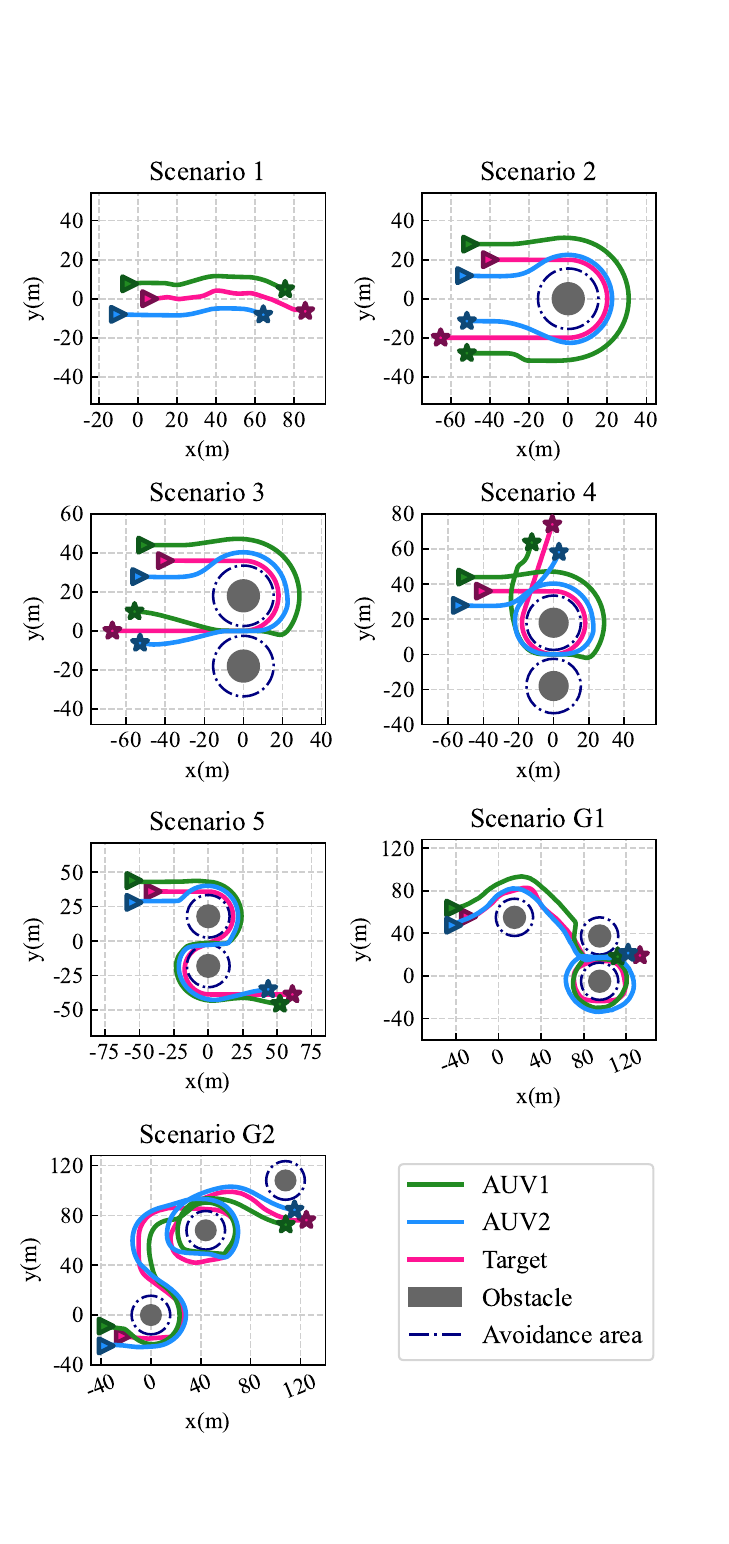}
  \caption{Trajectories of the target and AUVs of the expert demonstrations, and obstacle distributions in different featured scenarios. }
  \label{fig_4}
  \end{figure}

As traditional control methods and reward function-based RL methods may not viable in multi-objective tasks, we propose the efficient training framework FISHER, which utilizes expert demonstrations. The overall architecture of FISHER is depicted in Fig. \ref{fig_2}, and the pseudo-code refers to Algorithm 1 and Algorithm 2. Above all, the sim2sim procedure is first executed to accomplish the domain transfer of demonstrations and collect the expert replay buffer. Then in the first stage, MADAC is utilized for independent training and trajectory collecting in its corresponding demonstration scenario, which can be executed in parallel, and all trajectories are consolidated to generate a large-scale 
offline dataset that guarantees the capability of ORL. Subsequently, in the second stage, MAIGDT leverages the HIM transformer $\Phi$ to learn to approximate trajectories in the offline dataset sufficiently. Finally, policies applicable across various scenarios with one-shot demonstration can be acquired.

To make it clear, our proposed FISHER distinguishes itself from existing IL+DRL approaches through three key innovations:
$1^\circ$ Two-Stage Decoupled Framework—FISHER is the first to sequentially decouple imitation learning (MADAC stage) and reward-free offline RL (MAIGDT stage). The MADAC stage employs a Nash equilibrium-driven centralized discriminator to efficiently bootstrap policies, while the MAIGDT stage achieves generalization without manual reward engineering via latent state-matching. This eliminates the intertwined dependency issues seen in traditional methods.
$2^\circ$ Multi-Agent Specialized Design—A centralized discrimination-decentralized execution architecture enhances sample efficiency and scalability. The novel hindsight information matching technique enables collaborative learning solely through state transitions, overcoming limitations of shared reward assumptions in dense obstacle scenarios.
$3^\circ$ Sim2Sim Demonstration Generation—We pioneer an RL-based sim2sim pipeline to adaptively convert potential field-based expert trajectories from simplified environments into complex underwater demonstrations. This overcomes the challenges of costly real-world demonstrations and idealized simulation biases.

Collectively, these innovations enable FISHER to surpass existing methods in generalization capability, stability, and scalability.

\begin{figure}[!htb]
 \centering
  \subfloat[Ideal situation]{\centering\includegraphics[width=0.49\linewidth]{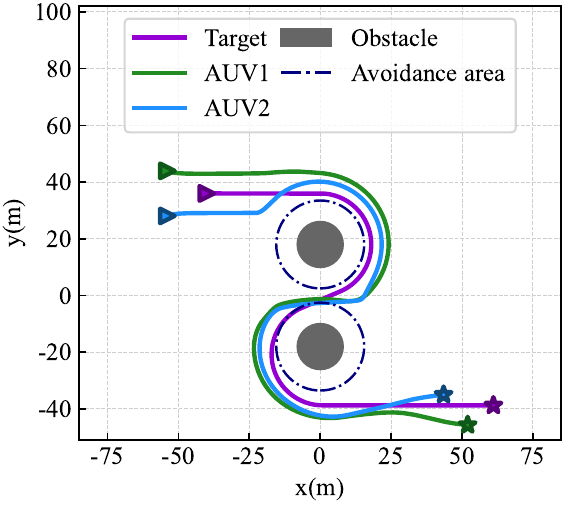}\label{fff1}}
  %\hfil
  \subfloat[An example with introduced random variations]{\centering\includegraphics[width=0.49\linewidth]{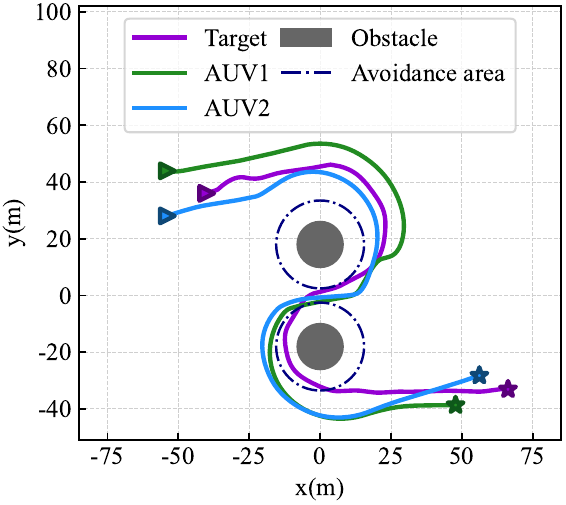}\label{fff2}}
  \caption{Trajectories of the target and AUVs of the expert demonstrations, taken from scenario 5. (a) Ideal situation. (b) An example with introduced random variations.}
  \end{figure}

% MADAC VS MAIDAC
\begin{figure}[!t]
  \centering
  \subfloat[MADAC ($N=2,3,4$)]{\centering\includegraphics[width=0.95\linewidth]{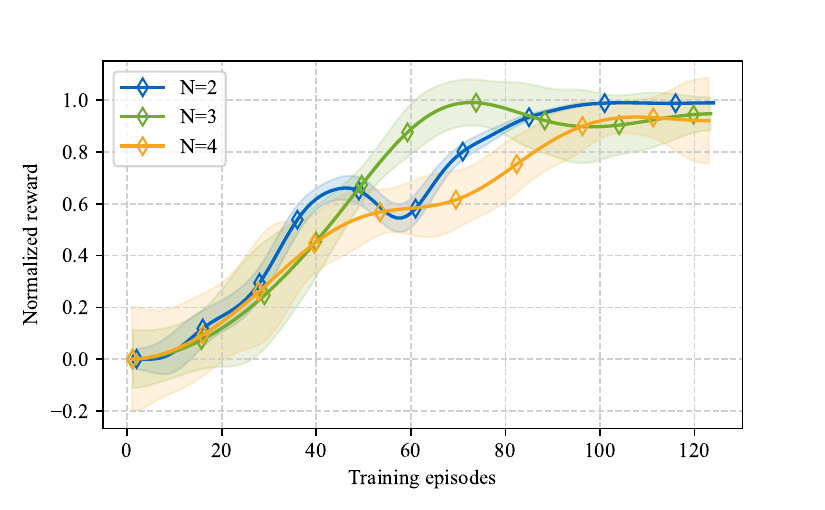}\label{fff1}%
  \label{}}\\
  %\hfil
  \centering
  \subfloat[MAIDAC ($N=2,3,4$) and GAIL+PPO ($N=2$)]{\centering\includegraphics[width=0.95\linewidth]{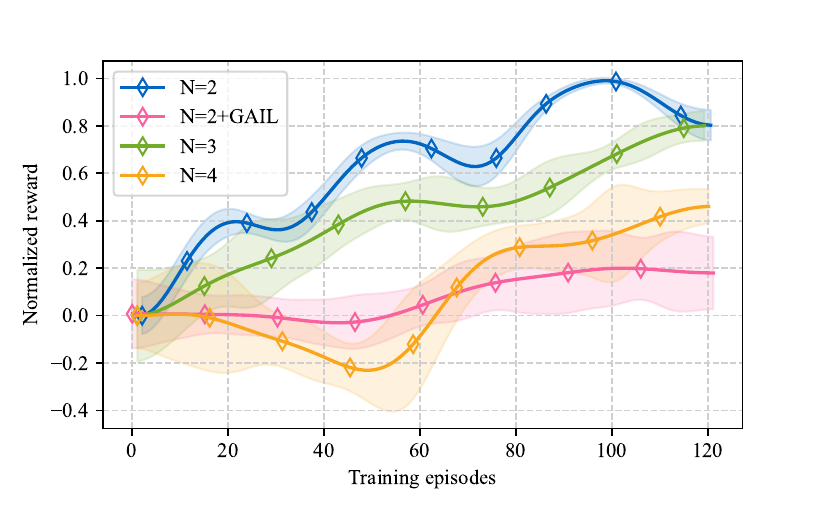}\label{fff2}%
  \label{}}
  \caption{The training curves of (a) MADAC ($N=2,3,4$). (b) MAIDAC ($N=2,3,4$) and GAIL+PPO ($N=2$).}
  \label{fig5}
  \end{figure}
  % FIG6 - 普通图像,后面看情况要不要移动一下
\begin{figure}[!t]
  \centering
  \includegraphics[width=0.95\linewidth]{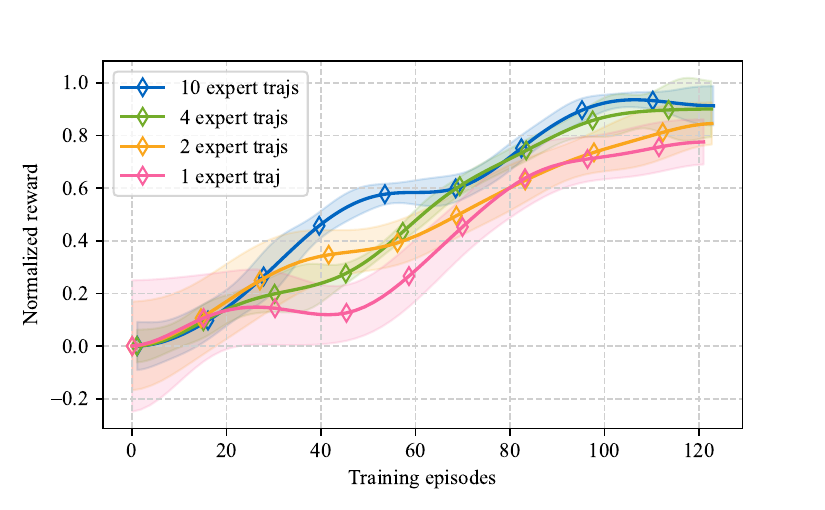}
  \caption{The training curves of MADAC given 10(default), 4, 2, 1 expert trajectories. }
  \label{fig_6}
  \end{figure}
\begin{figure*}[!t]
 \centering
  \subfloat[$N=2$]{\centering\includegraphics[width=0.315\linewidth]{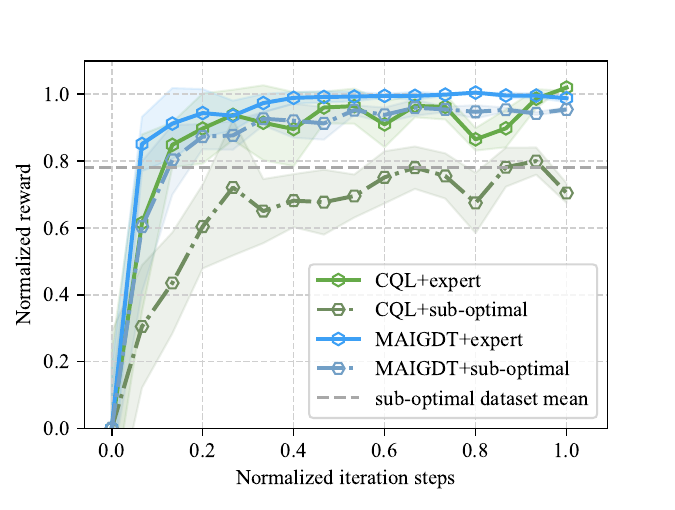}\label{fff1}}
  %\hfil
  \subfloat[$N=3$]{\centering\includegraphics[width=0.315\linewidth]{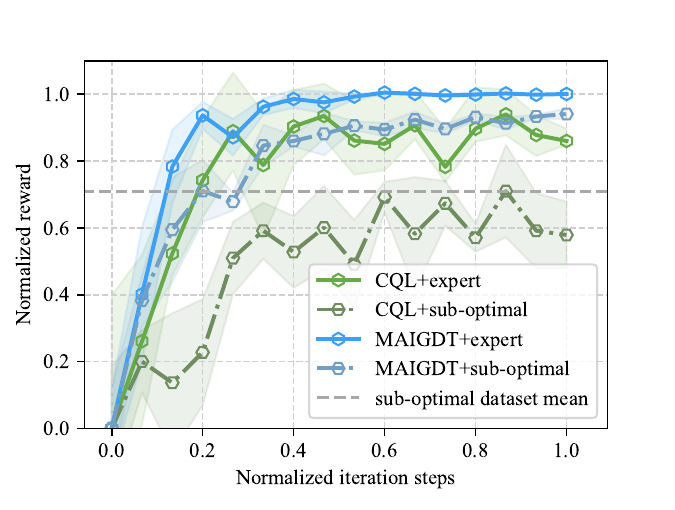}\label{fff2}}
    \subfloat[$N=4$]{\centering\includegraphics[width=0.315\linewidth]{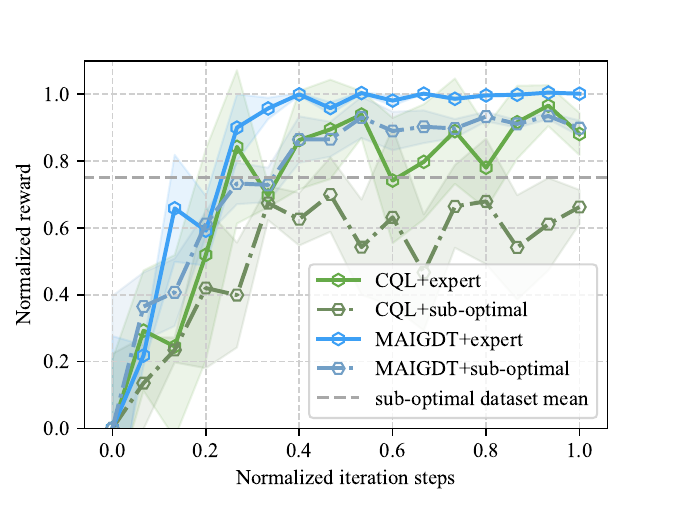}\label{fff3}} 
  \caption{The training curves of CQL and MAIGDT utilizing expert and sub-optimal dataset with (a) $N=2$ AUVs. (b) $N=3$ AUVs. (c)  $N=4$ AUVs.}
  \label{fig_7}
  \end{figure*}

\section{Simulation Results And Discussions} 

In this section, we demonstrate the performance of the FISHER framework through simulation experiments. First, we introduce the settings of the simulation experiments. Then, we detail the experiment scenario and demonstration design. Subsequently, we present the design of performance metrics, simulation results, and detailed discussions.

\subsection{Experiment Settings} % 奖励被放缩

% 注意：①实验关于距离等信息需要描述更详细 
To evaluate the effectiveness of FISHER in simulation, we first establish motion constraints and experimental configurations: while pure rotational motions (\(v=0, w \neq 0\)) are permitted, they are strictly bounded by \(w_{\text{max}}\) and dynamically validated in real-time against the AUV's physical model to ensure feasibility. In our experiments, all AUVs and the target are initialized facing the positive \(x\)-axis, with the target guaranteed to lie within each AUV's detection range—a setup enabled by the rotation-invariant state space design (Section III), which inherently decouples orientation dependencies. The AUVs operate at 12.5\,Hz to track the target moving at 1.2\,\text{m/s}, with their commanded velocities (\(v^{\text{des}}, \omega^{\text{des}}\)) dynamically regulated by the predefined constraints (\(v_{\max}, \omega_{\max}\)) and continuously verified through the embedded dynamic model. This integrated approach ensures both algorithmic validity and physical realizability.

% 对于算法参数设置，对比算法均采用了其原始论文进行设定。DAC的策略使用了SAC，相关的参数亦主要借鉴了SAC。MAIGDT的参数设置业主要参照GDT。
For algorithm parameters, since the MADAC utilize SAC\cite{26} as its policy, the related parameters are also mainly referenced from SAC. Similarly, the parameters setting of MAIGDT mainly refer to DT\cite{24}. In our implementation, all networks (actor/critic/discriminator) for both MADAC and SAC are realized as three-layer MLPs with hidden layers containing 128 units. For MAIGDT, both the policy transformer and HIM employ lightweight 3-layer decoder-only transformer architectures, featuring single attention heads and 128-dimensional embeddings. Additionally, for the baseline algorithms for comparison, the parameters are set according to the original paper. The other parameters of the environment and algorithm are mainly listed in Table I for summary.

% 分隔开

\subsection{Experiment Scenarios and Demonstrations}

Several scenarios are designed to evaluate the performance of proposed MADAC and MAIGDT algorithms employed in the overall FISHER framework. These scenarios feature different target moving trajectories and obstacle distributions, and we design the corresponding demonstrations for them. We divide these scenarios into two parts as shown in Fig. 5:

\begin{figure}[!t]
  \centering
  \subfloat[CQL ($N=2,4$)]{\centering\includegraphics[width=0.95\linewidth]{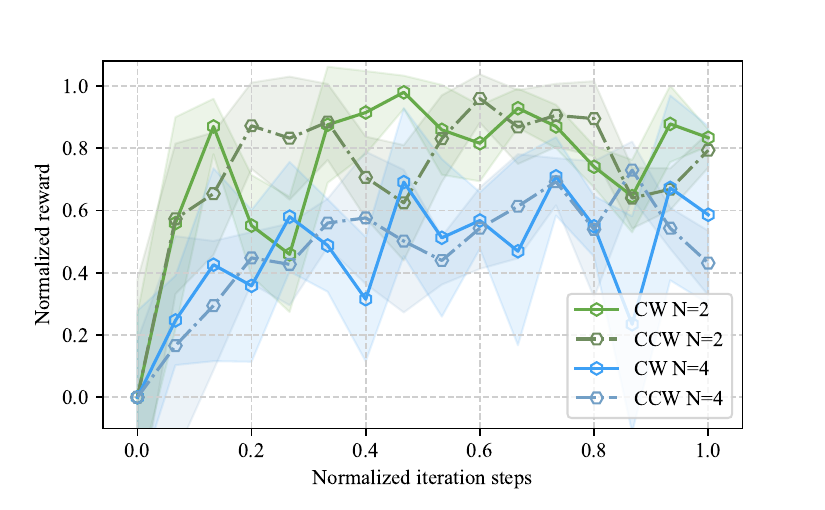}\label{fff1}%
  \label{}}\\
  %\hfil
  \centering
  \subfloat[MAIGDT ($N=2,4$)]{\centering\includegraphics[width=0.95\linewidth]{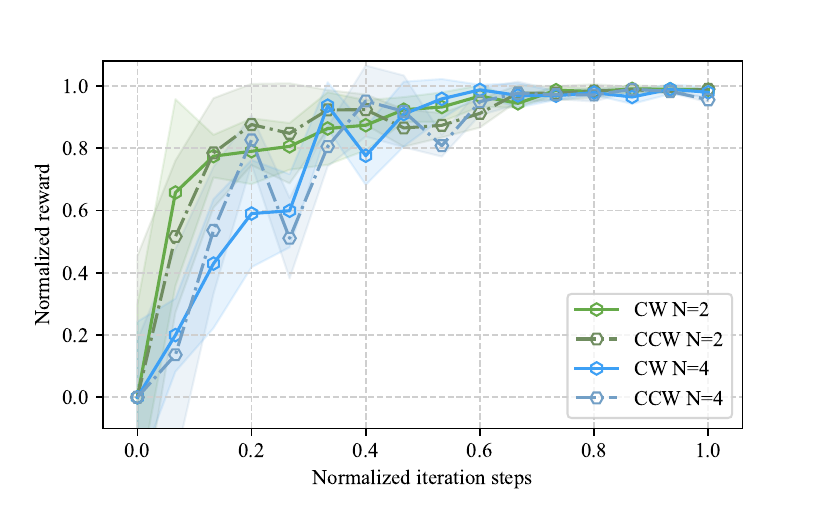}\label{fff2}%
  \label{}}
  \caption{The training curves of CW and CCW tasks taken from scenario 2. (a) Utilizing CQL ($N=2, 4$) for training. (b) Utilizing MAIGDT ($N=2, 4$) for training.}
  \label{fig8}
  \end{figure}

The first part features sparse obstacle(s). As all the objectives are relatively uncomplicated to be optimized, RL methods usually demonstrate passable performance in this part. Here, we design two scenarios, and we label them as scenario 1 and scenario 2.

In contrast, the second part, which features dense obstacles, presents a challenge to optimization given the reward function, as AUVs must weigh the objectives dynamically. For example, AUVs must reorganize their formation while passing through obstacles. Accordingly, we also design two scenarios, which are labeled as scenario 3 and scenario 4. 

To evaluate the generalization performance of FISHER, we introduced Scenarios G1 and G2 based on Scenarios 1-5. These new scenarios feature longer durations, non-fixed obstacle arrangements, target trajectories, and target speeds varying between 0.8 m/s and 1.5 m/s. Additionally, we ensure that the sim2sim and offline datasets are exclusively generated from Scenarios 1-5 to guarantee an accurate evaluation. The position of the target and obstacle(s) and expert trajectories of these mentioned scenarios are shown in Fig. 5 in detail.

To enhance the complexity of the diversity of the demonstrations, we implement some measures, which involve introducing random fluctuations to the trajectories of the target and demonstrations, add minor changes to the positions of obstacles, adjusting specific parameters (e.g., the radius of obstacles, target velocity) of the scenarios, and occasionally causing the target to deviate from its predefined trajectory. Fig. 6 shows an example with introduced random variations of Scenario 5. 

% 其将不会被纳入offline dataset里面。

% CAPTION: Trajectories of the target, expert demonstrations of AUVs ($N=2$) and obstacle distribution of Scenario 1 (sparse obstacle), Scenario 2 (sparse obstacle), Scenario 3 (dense obstacles) and Scenario 4 (dense obstacles).

\subsection{Experiment Results and Analysis}

Various experiments are conducted based on these scenarios mentioned before. Firstly, we evaluate the performance of MADAC and MAIGDT through comparative experiments in scenarios with sparse obstacle(s) quantitatively using the reward function, which is relatively capable of aligning with our demands in simple situations.

Fig. 7 displays the training results of MADAC, multi-agent DAC with a decentralized setting (MAIDAC), and a mainstream implementation of GAIL (GAIL+PPO), in scenario 1 with the number of AUVs ranging from $N=2$ to $N=4$. It should be noted that the reward between different $N$ cannot be compared directly. Consequently, we normalize the reward, such that the average reward obtained by randomly initialized policies is recorded as 0, while the reward from the expert trajectory is set to 1. Observations from Fig. 7(b) demonstrate that the GAIL method, due to its low sample efficiency and unsatisfactory training stability, shows virtually no policy improvement even in the simplest case of $N=2$. Although MAIDAC, which adopts a decentralized setting, shows little difference from MADAC at $N=2$, training becomes unstable from $N=3$ onwards, encountering a significant bottleneck, which indicates that AUVs can hardly explore advantageous states. In contrast, the training curve of MADAC for each number $N$ is stable, and finally, MADAC achieves a reward close to that of experts. 

Moveover, Fig. 8 illustrates the training curves of MADAC given a different number of demonstrations when $N=4$. As illustrated in the results, with the increase of the demonstrations (expert trajectories), the normalized reward shows an upward trend at the same training epoch, which showcases the acceleration and improvement effects on the training process. Besides, despite the presence of a reduction of performance for fewer demonstrations, MADAC does not necessitate an excessive number of expert trajectories, thereby ensuring sufficient training stability.
\begin{figure*}[!t]
 \centering
  \subfloat[Cooperative $N=2$]{\centering\includegraphics[width=0.303\linewidth]{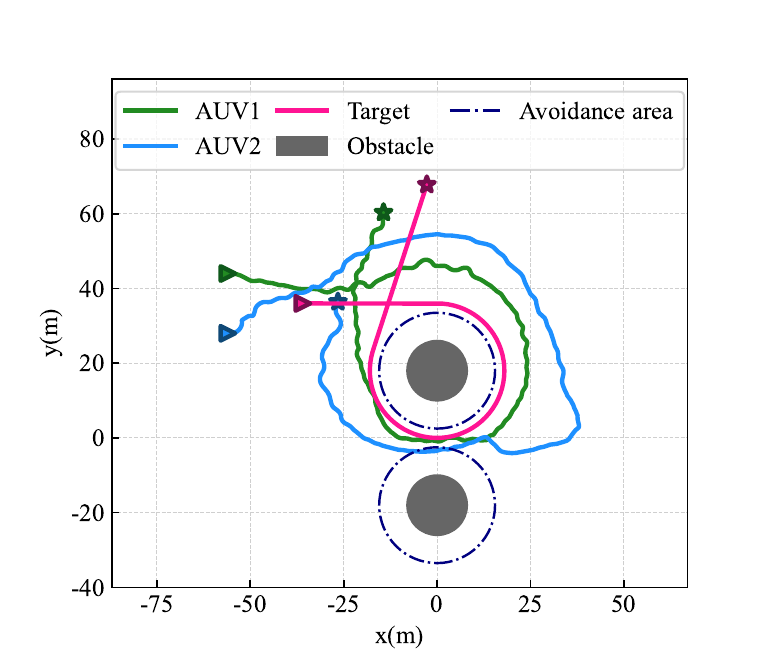}\label{fff1}}
  %\hfil
  \subfloat[Mixed $N=2$]{\centering\includegraphics[width=0.303\linewidth]{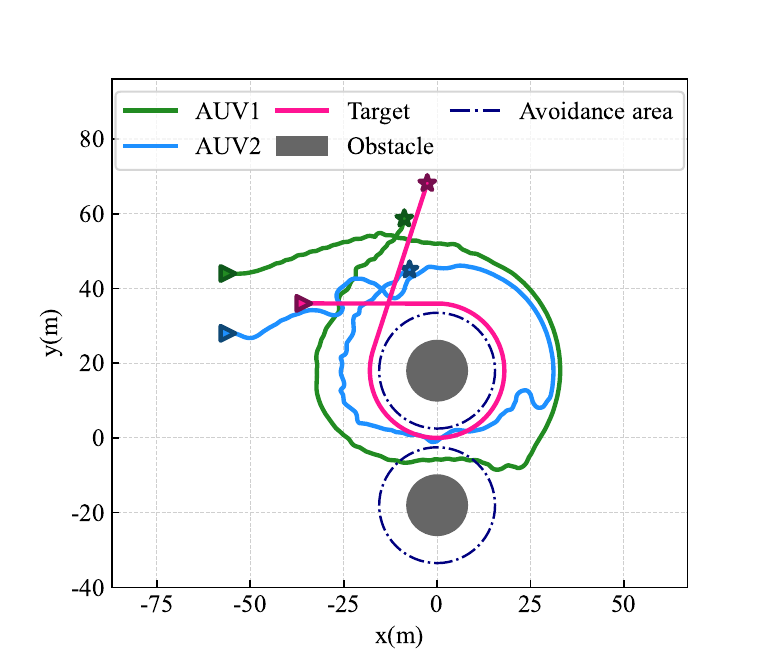}\label{fff2}}
    \subfloat[Split $N=2$]{\centering\includegraphics[width=0.303\linewidth]{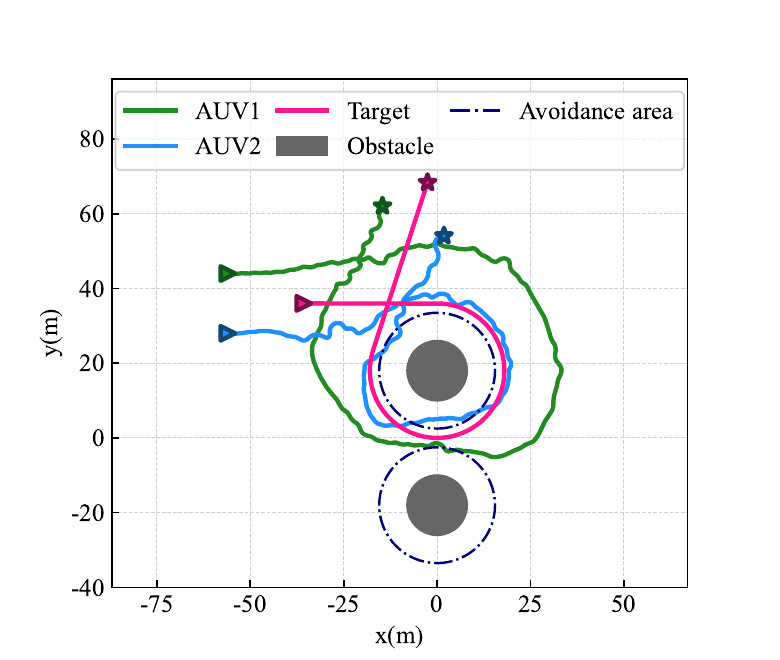}\label{fff3}} \\
    \subfloat[FISHER $N=2$]{\centering\includegraphics[width=0.303\linewidth]{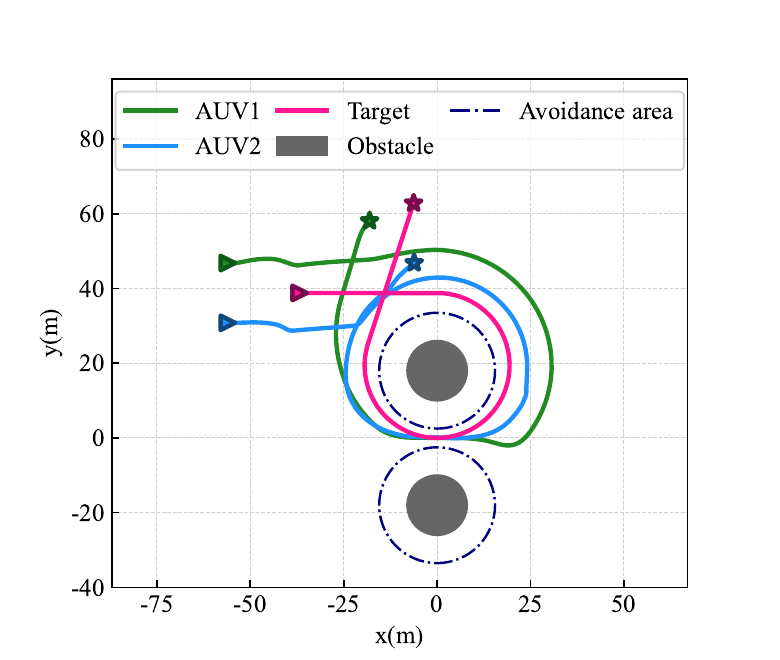}\label{fff4}}
    \subfloat[FISHER $N=3$]{\centering\includegraphics[width=0.303\linewidth]{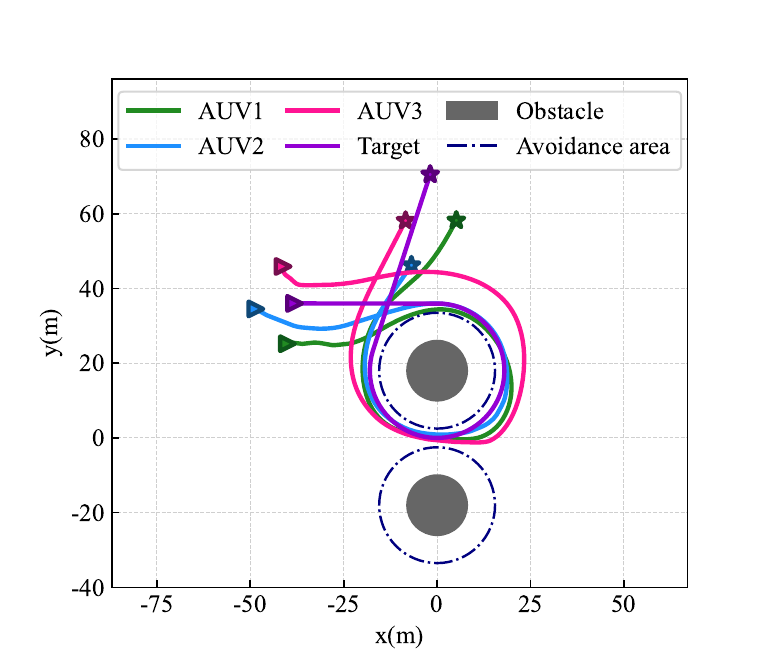}\label{fff5}}
    \subfloat[FISHER $N=4$]{\centering\includegraphics[width=0.30\linewidth]{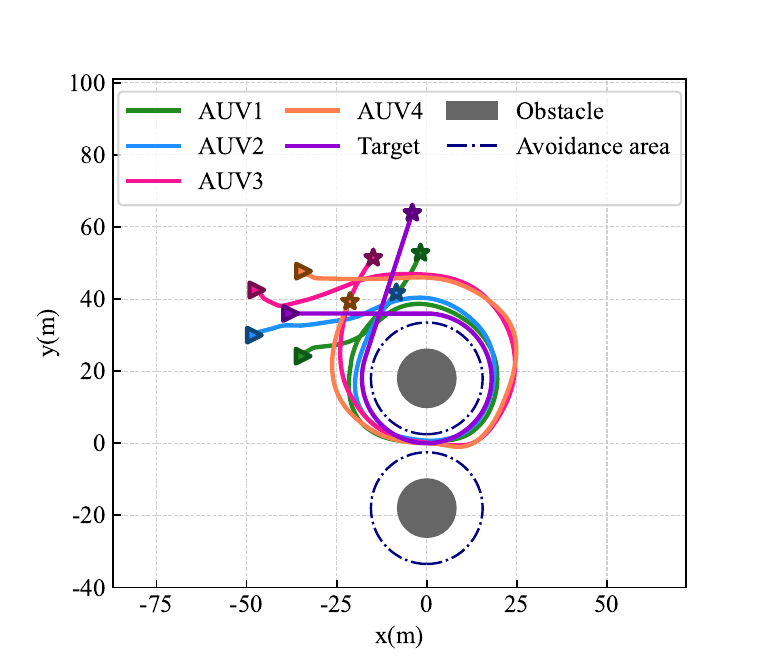}\label{fff6}}
  \caption{Representative tracking trajectories of AUVs via FISHER and SAC+CTDE under different reward settings and AUV number, respectively. (a) SAC+CTDE+Cooperative.  (b) SAC+CTDE+Mixed. (c) SAC+CTDE+Split. (d) FISHER ($N=2$). (e) FISHER ($N=3$). (f) FISHER ($N=4$).}
  \label{fig_10}
  \end{figure*}

\begin{figure}[!t]
  \centering
  \includegraphics[width=0.74\linewidth]{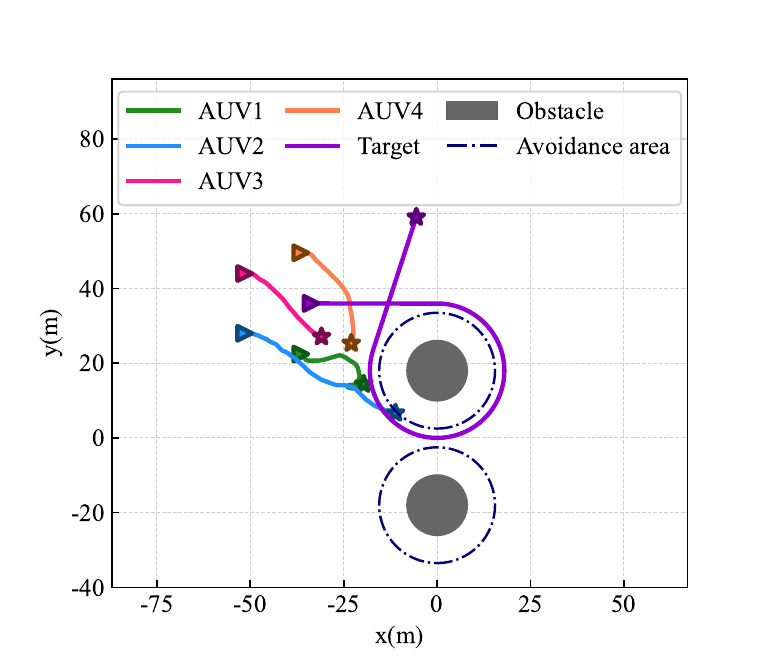}
  \caption{An example of tracking failure utilizing SAC+CTDE for training with $N=4$ AUVs. }
  \label{fig_11}
  \end{figure}

\begin{table*}
  \centering
  \caption{Performance Metrics of AUVs Tracking Target Utilizing SAC+CTDE and Proposed FISHER Framework in Scenario 4. } 
  \label{tab:feature-splitting}
  \resizebox{.90\linewidth}{!}{
    \begin{tabular}{l|cccccc}
      \toprule\!
      \textbf{Experiments} &  Cooperative & Mixed &  Split & FISHER &  FISHER & FISHER  \\
      &  $N\!=\!2$ & $N\!=\!2$ &  $N\!=\!2$ & $N\!=\!2$ &  $N\!=\!3$ & $N\!=\!4$ \\
      \midrule
      $\mathbb{E}$(min-distance) /m & $16.38\pm 0.30$ & $15.14\pm 0.89$& $14.16\pm 0.79$& $13.02\pm 0.89$& $12.50\pm 1.13$ & $12.24\pm 1.37$ \\
      Std(min-distance) /m & $2.95\pm 0.55$ & $2.61\pm 0.41$& $2.38\pm 0.59$& $2.02\pm 0.35$& $2.30\pm 0.56$ & $2.57\pm 0.64$ \\
      $\mathbb{E}$(consistency) & $87.69 \pm 4.44$ &  $96.11 \pm 1.94$& $97.88 \pm 0.91$& $97.53 \pm 0.69$& $140.99 \pm 3.55$ & $191.12 \pm5.09$  \\
      Std(consistency) & $5.35 \pm 1.42$&  $4.88 \pm 1.58$ & $2.99 \pm 0.64$& $1.42 \pm 0.37$ & $4.97 \pm 2.60$ &  $8.20 \pm 3.98$  \\
      Min(obs-distance) /m& $7.92 \pm 0.76$ & $6.16 \pm 0.66$ & $3.91 \pm 0.79$& $10.28 \pm 0.15$& $9.19 \pm 0.87$& $9.01 \pm 0.74$ \\
      Danger time /s& $8.44 \pm 2.13$  &$12.23 \pm 3.17$ & $16.64 \pm 4.65$ & $0.00 \pm 0.00$& $0.00 \pm 0.00$ &$0.00 \pm 0.00$  \\
      \bottomrule
  \end{tabular}}
\end{table*}

On the other hand, to verify the superior performance of MAIGDT, we conduct the comparative experiments in scenario 1 utilizing MAIGDT and classical TD-based ORL algorithm consevative Q-learning (CQL)\cite{27}, employing the expert dataset from MADAC (expert) and the dataset with sub-optimal trajectories from SAC (sub-optimal), respectively. The training results of MAIGDT and CQL are depicted in Fig. 9. The mean and standard deviation of the sub-optimal dataset trajectory rewards are represented in the figure with dashed lines and semi-transparent fill. Although the training curves of CQL also exhibit an upward trend, capable of enabling policy improvement. Nevertheless, significant fluctuations persist in the training process even after convergence. This is unacceptable, due to the unpredictable performance in the policy deployment. Conversely, MAIGDT’s regression-based training approach ensures training stability. Furthermore, in situations of dataset degradation and an increase in the AUV number $N$, CQL’s performance drastically deteriorates, while GDT, by learning the state transition rather than explicit rewards, can still adeptly replicate the performance of expert demonstrations.

Furthermore, we conduct extensive experiments to assess the multi-task performance of MAIGDT. To accomplish this, we formulate two tasks, both originating from scenario 2, but with forward directions rotating clockwise (CW) and counterclockwise (CCW). The results of the experiment are illustrated in Fig. 10. For $N=2$, the training outcomes of CQL are quite unstable, with the reward of the two tasks exhibiting significant fluctuation. For $N=4$, CQL is incapable of reaching convergence, as the reward function fails to align with the optimization objective of the target tracking task, which is further demonstrated later in this section. Still, MAIGDT can achieve the best performance in both tasks.

% 图10尝试绘制，由于目前暂时没有完成图像的绘制，因此就暂时全部用一张图片替代
% subplot的Caption首字母大写

% 由于引入了N=3/4，不知道该怎么给文本和数据加粗

Subsequently, we further evaluate the overall performance of the FISHER framework in scenarios with dense obstacles. As the environment becomes more complicated, the reward function is unsuitable for evaluating performance. For quantitative evaluation, we introduce some performance indicators similar to Yang \MakeLowercase{\textit{et al.}}\cite{17}: minimum distance mean, minimum distance standard deviation, consistency mean, consistency standard derivation, minimum obstacle distance and danger time. Here, the minimum distance represents the distance between the target and the nearest AUV to the target, the minimum obstacle distance refers to the shortest distance between the obstacle and the AUVs throughout the entire process, while the duration in danger is defined as the period which at least one AUV that is less than 8m away from an obstacle, and consistency is represented by the algebraic connectivity $\lambda$. Due to the randomness of RL training process, we train the policies until convergence from scratch 3 times to measure the training stability, and all results are presented as $a\pm b$, where $b$ is the standard deviation of metrics between these policies.

\begin{figure*}[!t]
 \centering
    \subfloat[Scenario G2, $N=2$]{\centering\includegraphics[width=0.305\linewidth]{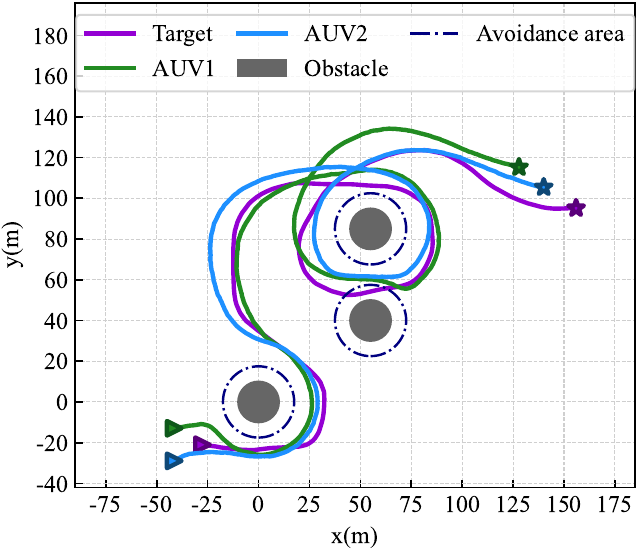}\label{fff4}}
    \hspace{1.5mm}
    \subfloat[Scenario G2, $N=3$]{\centering\includegraphics[width=0.302\linewidth]{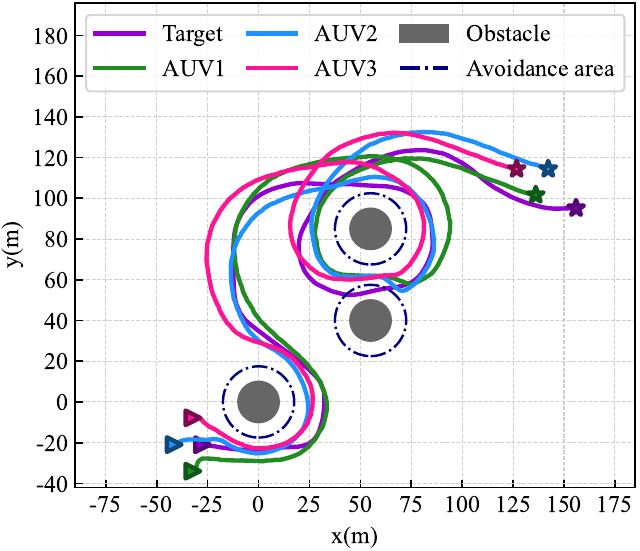}\label{fff5}}
    \hspace{1.5mm}
    \subfloat[Scenario G2, $N=4$]{\centering\includegraphics[width=0.305\linewidth]{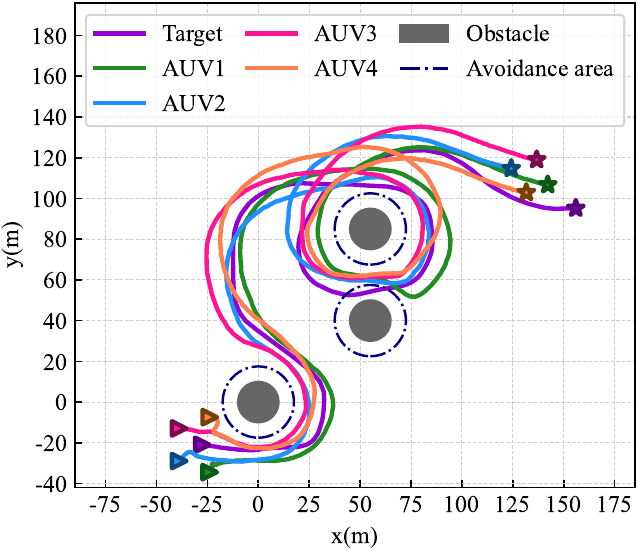}\label{fff6}}
  \caption{Representative tracking trajectories of AUVs via FISHER with different AUV number, taken from scenario G2. (a) Scenario G2, $N=2$. (b) Scenario G2, $N=3$. (c) Scenario G2, $N=4$.}
  \label{fig_10}
  \end{figure*}

\begin{table*}
  \centering
  \caption{Performance metrics of AUVs tracking target utilizing proposed FISHER framework in scenario G1 and G2.} 
  \label{tab:feature-splitting}
  \resizebox{.90\linewidth}{!}{
    \begin{tabular}{l|cccccc}
      \toprule\!
      \textbf{Experiments} &  Scenario G1 & Scenario G1 & Scenario G1 & Scenario G2 &  Scenario G2 & Scenario G2  \\
      &  $N\!=\!2$ & $N\!=\!3$ &  $N\!=\!4$ & $N\!=\!2$ &  $N\!=\!3$ & $N\!=\!4$ \\
      \midrule
      $\mathbb{E}$(min-distance) /m & $12.56\pm 0.47$ & $12.36\pm 0.61$& $11.93\pm 0.66$& $14.51\pm 1.18$& $13.09\pm 1.50$ & $12.69\pm 1.62$ \\
      Std(min-distance) /m & $1.39\pm 0.40$ & $1.17\pm 0.41$& $1.14\pm 0.73$& $1.84\pm 0.71$& $1.89\pm 0.77$ & $2.03\pm 1.12$ \\
      $\mathbb{E}$(consistency) & $99.10 \pm 1.18$ &  $147.42 \pm 2.39$& $196.37 \pm 2.94$& $97.10 \pm 2.45$& $138.48 \pm 3.74$ & $187.69 \pm4.80$  \\
      Std(consistency) & $1.13 \pm 0.62$&  $1.67 \pm 0.90$ & $2.41 \pm 1.11$& $2.54 \pm 0.43$ & $5.32 \pm 2.31$ &  $7.71 \pm 3.85$  \\
      Min(obs-distance) /m& $13.36 \pm 2.21$ & $10.88 \pm 1.69$ & $9.28 \pm 1.82$& $9.66 \pm 0.68$& $8.82 \pm 0.47$& $8.40 \pm 0.53$ \\
      Danger time /s& $0.00 \pm 0.00$  &$0.00 \pm 0.00$ & $0.00 \pm 0.00$ & $0.00 \pm 0.00$& $0.00 \pm 0.00$ &$1.06 \pm 1.22$  \\
      \bottomrule
  \end{tabular}}
\end{table*}

Additionally, we introduce a MADRL-based baseline to reveal critical challenges in reward engineering. Specifically, we adapt the SAC algorithm \cite{26} for continuous action domains with the integration of centralized training procedures with decentralized execution (CTDE) manner (SAC+CTDE). The SAC+CTDE baseline is trained with three settings of the reward function mentioned in Section \uppercase\expandafter{\romannumeral3}, namely cooperative, mixed and split, respectively. Performance metrics of SAC+CTDE with three reward settings in $N=2$ and MAIGDT in $N=2,3,4$ of scenario 4 are shown in Table \uppercase\expandafter{\romannumeral2}, and the corresponding trajectories are shown in Fig. 11. Additionally, the minimum distance mean and minimal obstacle distance of expert demonstrations are 12m and 10.8m, respectively, while the consistency are $100.1$ for $N=2$, $150.2$ for $N=3$, and $200.2$ for $N=4$. 

% show a failed example

For $N=2$, SAC+CTDE performs passable capability only in certain performance metrics, such as better tracking performance in the split setting and better obstacle avoidance in the cooperative setting. However, all reward functions fail to achieve a balance under multiple objectives. The trajectories of AUVs are not smooth and exhibit significant jitters. Moreover, SAC+CTDE fails to track the target and AUVs turn back when encountering obstacles in $N=3$ and $N=4$ with any reward function of the three, due to the increasing deviation of the reward function from expected optimization goal. The representative example of failed tracking processes of SAC+CTDE is shown in Fig. 12. Conversely, FISHER successfully replicates demonstrations, showcasing superior performance comparable to those of experts, while ensuring stability as the number of AUVs increases.

Finally, we further test the generalization performance in Scenarios G1 and G2. The tracking trajectories are illustrated in Fig. 13, and the corresponding performance metrics are presented in Table \uppercase\expandafter{\romannumeral2}. These metrics are statistically derived from 10 evaluations with randomized target trajectories and obstacle positions. The results demonstrate that FISHER exhibits strong robustness and maintains expert-level performance.

% 放在最后去
\subsection{Potential Real-World Applications}
The FISHER framework, while initially validated in 2D simulations, enables several important real-world marine applications through its innovative approach to multi-AUV coordination.  The system's robust performance in maintaining formation integrity while avoiding obstacles (Section V) makes it particularly valuable for underwater infrastructure inspection, where AUV teams must navigate complex environments like offshore wind farms or pipeline networks while maintaining precise relative positioning.  The framework's ability to adapt to dynamic targets (Figs. 11-13) directly supports search and rescue operations in ocean currents, where multiple vehicles must cooperatively track moving objects or persons while compensating for hydrodynamic disturbances.

For marine scientific research, FISHER's distributed coordination capabilities enable efficient large-area monitoring of coral reefs or algal blooms, where its communication-range-aware design (Section III-C) ensures continuous data collection even in challenging acoustic conditions.  The system's obstacle avoidance performance (Table II) is particularly relevant for operations in cluttered environments like shipwrecks or underwater caves, where conventional single-AUV approaches face higher collision risks.

% By contrast, FISHER still successfully maneuvers AUVs to follow the target while strictly obeying the constraints of safe distance. 

\section{conclusion and Discussion} % conclusion也应该是现在时
In this paper, we developed FISHER, an efficient training framework that leverages expert demonstrations generated from sim2sim for multi-AUV underwater target tracking task. We first introduced DAC to enhance the sample efficiency and training stability of the GAIL-based algorithm, and we expanded it to MADAC by optimizing the dual problem with the Nash equilibrium constraint. Then, MAIGDT was introduced to attain multi-task applicable policies with the help of latent variables of demonstrations from the anti-casual information extractor rather than designing reward functions like DT and other ORL methods. The MADAC and MAIGDT together constitute the two stages of FISHER framework. Simulation results in multiple scenarios reveal that FISHER excellently learns from demonstrations and achieves superior performance levels comparable to expert trajectories. 

Building on our current 2D framework, we plan to systematically progress toward real-world 3D deployment through a series of targeted enhancements. Our first step involves extending the state-action space to incorporate depth dynamics and volumetric perception, thereby enabling applications such as underwater structure inspection and full-volume ocean monitoring. Following this, we will assess system robustness in increasingly complex conditions, including turbulent hydrodynamic environments as well as dynamic scenarios involving moving obstacles and varying visibility. These evaluations will be conducted using high-fidelity simulations built upon our established turbulence modeling framework.

To bridge the sim-to-real gap, we will construct a comprehensive transfer pipeline that integrates hardware-in-the-loop (HIL) validation with real AUV platforms, applies domain randomization to account for uncertainties in sensing and actuation, and leverages our sim2sim methodology to progressively minimize the reality gap. Collectively, these developments will support the reliable deployment of our system in practical marine settings, while retaining its core capabilities in multi-agent coordination.

In parallel, we are actively integrating the FISHER framework into embedded AUV hardware and validating its real-world performance through both tank tests and open-water trials under dynamic environmental disturbances. Thanks to its lightweight and modular design, FISHER supports real-time execution on resource-constrained platforms such as Jetson NX. Moreover, its reward-free and task-general architecture enables flexible multi-agent coordination without the need for retraining. Together, these efforts lay foundation for scalable and robust deployment in real-world marine environments.

% \section*{Acknowledgments}
% This should be a simple paragraph before the References to thank those individuals and institutions who have supported your work on this article. 

% NOTE 注意严格检查符号的规范性

{\appendix[Proof to The Training Objecive of MADAC]
Hereinafter, $\hat{v}_i(\boldsymbol{s})$, $\hat{q}_i(\boldsymbol{s},a_i)$ represent $\hat{v}_i(\boldsymbol{s};\boldsymbol{\pi},\boldsymbol{r})$ and $\hat{q}_i(\boldsymbol{s},a_i;\boldsymbol{\pi},\boldsymbol{r})$, respectively. 

 \textbf{\textit{Lemma 1:}} For the value function $\hat{v}_i(\boldsymbol{s})$ that meets the condition of the Bellman equation
 \begin{equation}
  \hat{v}_i(\boldsymbol{s}) = \mathbb{E}_{\boldsymbol{\pi}}[r_i(\boldsymbol{s},\boldsymbol{a})+\gamma\sum_{\boldsymbol{s}'\in\boldsymbol{S}}\boldsymbol{P}(\boldsymbol{s}'|\boldsymbol{s},\boldsymbol{a})\hat{v}_i(\boldsymbol{s}')].
 \end{equation}
 
 Then $\hat{q}_i(\boldsymbol{s},a_i)=\mathbb{E}_{\pi_{-i}}[r_i(\boldsymbol{s},\boldsymbol{a})+\gamma\sum_{\boldsymbol{s}'\in\boldsymbol{S}}\boldsymbol{P}(\boldsymbol{s}'|\boldsymbol{s},\boldsymbol{a})\hat{v}_i(\boldsymbol{s}')]$ is defined similarly, where $\pi_{-i}$ denotes all policies except the policy of $i$-th AUV. Then we can obtain
 \begin{itemize}
  \item{$1^{\circ}$ For any $\boldsymbol{\pi}$, $f_{\boldsymbol{r}}(\boldsymbol{\pi},\boldsymbol{v}) = 0.$} 
  \item{$2^{\circ}$ $\boldsymbol{\pi}$ is the Nash equilibrium under $\boldsymbol{r}$ if and only if $\hat{v}_i(\boldsymbol{s}) \geq \hat{q}_i(\boldsymbol{s},a_i),\forall i \in \{1,\cdots,N\}.$} 
 \end{itemize}

 \textbf{\textit{Proof 1:}} By the definition of $\hat{v}_i(\boldsymbol{s})$, as actions are mutually independent conditioned on $\boldsymbol{s}$, we can obtain
 \begin{equation}
 \begin{aligned}
  \hat{v}_i(\boldsymbol{s}) & = \mathbb{E}_{\boldsymbol{\pi}}[r_i(\boldsymbol{s},\boldsymbol{a})+\gamma\sum_{\boldsymbol{s}'\in\boldsymbol{S}}\boldsymbol{P}(\boldsymbol{s}'|\boldsymbol{s},\boldsymbol{a})\hat{v}_i(\boldsymbol{s}')] \\
          & = \mathbb{E}_{{\pi_i}}[\mathbb{E}_{{\pi_{-i}}}[r_i(\boldsymbol{s},\boldsymbol{a})+\gamma\sum_{\boldsymbol{s}'\in\boldsymbol{S}}\boldsymbol{P}(\boldsymbol{s}'|\boldsymbol{s},\boldsymbol{a})\hat{v}_i(\boldsymbol{s}')]] \\
          & = \mathbb{E}_{{\pi_i}}[\hat{q}_i(\boldsymbol{s},a_i)].
\end{aligned}
\end{equation}

Therefore $1^\circ$ can be proved. For $2^\circ$, clearly the  Nash equilibrium conditions are violated if there exists $i \in \{1,\cdots,N\}$, $\boldsymbol{s}$ and $a_i$ that $\hat{v}_i(\boldsymbol{s}) < \hat{q}_i(\boldsymbol{s},a_i)$, namely the $i$-th AUV can choose $a_i$, when the corrsponding state is $\boldsymbol{s}$ and the policy follow $\pi_i$ subsequently, to achieve higher expected return. If the constraint is met, then
\begin{equation}
  \label{wow1}
  \hat{v}_i(\boldsymbol{s}) \geq \mathbb{E}_{\pi_i}[\hat{q}_i(\boldsymbol{s},a_i)] = \hat{v}_i(\boldsymbol{s}).
\end{equation}

The Eq. \eqref{wow1} signifies that when the constraint is met, there must only one solution of $\hat{v}_i(\boldsymbol{s})$. Hence, $2^\circ$ is proved.$\hfill\blacksquare$ % Consequently也是一个可行替代

Then we attempt to obtain the multiple-timestep TD equivalent of constraint:

\textbf{\textit{Theorem 2:}} Assume that the AUVs' trajectory from timestep $0$ to $t-1$ is denotes as $\{\boldsymbol{s}^{(j)},\boldsymbol{a}^{(j)}\}^{t-1}_{j=0}$, then the state for timestep $t$ is $\boldsymbol{s}^{(t)}$, and the action of $i$-th AUV is $a_i^{(t)}$, we denote the discounted expected return as
\begin{equation}
\begin{aligned}
  & \hat{q}_i^{(t)}(\{\boldsymbol{s}^{(j)},\boldsymbol{a}^{(j)}\}^{t-1}_{j=0},\boldsymbol{s}^{(t)},a_i^{(t)}) \\
  = & \sum_{j=0}^{t-1}\gamma^j r_i(\boldsymbol{s}^{(j)},\boldsymbol{a}^{(j)}) \\
  + & \gamma^t\mathbb{E}[r_i(\boldsymbol{s}^{(t)},\boldsymbol{a}^{(t)})+\gamma\sum_{\boldsymbol{s}'\in\boldsymbol{S}}P(\boldsymbol{s}'|\boldsymbol{s},\boldsymbol{a}^{(t)})\hat{v}_i(\boldsymbol{s}')].
\end{aligned}
\end{equation}

Then $\boldsymbol{\pi}$ reaches a Nash equilibrium if and only if
\begin{equation}
  \begin{aligned}
  & \hat{v}_{i}(\boldsymbol{s}^{(0)})\geq\mathbb{E}_{\pi_{-i}}\left[\hat{q}_{i}^{(t)}(\{\boldsymbol{s}^{(j)},\boldsymbol{a}^{(j)}\}_{j=0}^{t-1},\boldsymbol{s}^{(t)},a_{i}^{(t)})\right] \\
  & \triangleq Q_{i}^{(t)}(\{\boldsymbol{s}^{(j)},a_{i}^{(j)}\}_{j=0}^{t}) , \forall t\in\mathbb{N}^{+},i\in\{1,\cdots,N\}.
  \end{aligned}
\end{equation}

\textbf{\textit{Proof 2:}} Similarly, we consider that the constraint does not comply, namely exists $i \in \{1,\cdots,N\}$, $\{\boldsymbol{s}^{(j)},\boldsymbol{a}^{(j)}\}_{j=0}^{t-1}$ that
\begin{equation}
  \hat{v}_i(\boldsymbol{s}^{(0)})<\mathbb{E}_{\pi_{-i}}[\hat{q}_i^{(t)}(\{\boldsymbol{s}^{(j)},\boldsymbol{a}^{(j)}\}_{j=0}^{t-1},\boldsymbol{s}^{(t)},a_i^{(t)})],
\end{equation}
then $i$-th AUV can achieve a higher expected return by choosing $a_i^{(j)}$ when correspond state is $\boldsymbol{s}^{(j)}$ and following $\pi_i$ subsequently. This contradicts the Nash equilibrium. If the constraint is met, then for all $i$ and trajectory $\{\boldsymbol{s}^{(j)},\boldsymbol{a}^{(j)}\}_{j=0}^{t-1}$, 
\begin{equation}
  \hat{v}_i(\boldsymbol{s}^{(0)}) \geq \mathbb{E}_{\pi_{-i}}[\hat{q}_i^{(t)}(\{\boldsymbol{s}^{(j)},\boldsymbol{a}^{(j)}\}_{j=0}^{t-1},\boldsymbol{s}^{(t)},a_i^{(t)})].
\end{equation}

As we can construct any $\hat{q}_i(\boldsymbol{s}^{(0)},a_i^{(0)})$, which has the equivalent
\begin{equation}
  \begin{aligned}
  & \hat{q}_i(\boldsymbol{s}^{(0)},a_i^{(0)}) \\
  & =\mathbb{E}_{\boldsymbol{\pi}}[\hat{q}_i^{(t)}(\{\boldsymbol{s}^{(j)},\boldsymbol{a}^{(j)}\}_{j=0}^{t-1},\boldsymbol{s}^{(t)},a_i^{(t)})] \\
  & =\mathbb{E}_{\pi_i}[\mathbb{E}_{\pi_{-i}}[\hat{q}_i^{(t)}(\{\boldsymbol{s}^{(j)},\boldsymbol{a}^{(j)}\}_{j=0}^{t-1},\boldsymbol{s}^{(t)},a_i^{(t)})]],
  \end{aligned}
\end{equation}
which is direct, as taking expectation over $\pi_i$ and $\pi_{-i}$ simultaneously takes it over states($\pi_{-i}$) and actions($\pi_i$). As the $\boldsymbol{s}^{(0)}$ and $a_i^{(0)}$ can be arbitrary, we can extend it to
\begin{equation}
  \hat{v}_i(\boldsymbol{s}) \geq \hat{q}_i(\boldsymbol{s},a_i).
\end{equation}

Then Theorem 1 can be proved according to Lemma 1. $\hfill\blacksquare$

According to Theorem 1, the optimizing objective of Nash equilibrium is always zero for the final solution. Therefore we can solve the dual problem of MARL and MAIRL by constructing the Lagrange multiplier
\begin{equation}
  \label{ldf}
\begin{aligned}
  \max_{\lambda\geq0}\min_{\boldsymbol{\pi}} \sum_{i=1}^{N}&\sum_{\tau_{i}\in\mathcal{T}_{i}^{t}}\lambda(\tau_{i})\left(Q_{i}^{(t)}(\tau_{i})-\hat{v}_{i}(\boldsymbol{s}^{(0)})\right) \\
  &  \triangleq  L_{\mathbf{r}}^{(t+1)}(\pi,\lambda),
\end{aligned}
\end{equation}
where $\mathcal{T}_{i}^{t}$ is the set of all possible $t$-timestep length sequence $\{\boldsymbol{s}^{(j)},\boldsymbol{a}^{(j)}\}_{j=0}^{t-1}$, $\boldsymbol{s}^{(0)} \sim \boldsymbol{P}_0(\boldsymbol{s})$ is the initial state. $\lambda$ is the vector of $N \cdot |\mathcal{T}_{i}^{t}|$ Lagrange multipliers, where $|\mathcal{T}_{i}^{t}|$ is the number of sequences in $\mathcal{T}_{i}^{t}$.

\textbf{\textit{Theorem 2:}} For any two sets of policies $\boldsymbol{\pi}$ and $\boldsymbol{\pi}'$, we define that the probability of generating the sequence $\tau_i$ with policy $\pi_i$ and $\pi_{-i}'$, namely
\begin{equation}
  \label{mtl}
  \begin{aligned}
    & \lambda_{\boldsymbol{\pi}}'(\tau_{i})=\boldsymbol{P}_0(\boldsymbol{s})\pi_{i}(a_{i}^{(0)}|\boldsymbol{s}^{(0)}) \\ 
    & \prod_{j=1}^{t}\pi_{i}(a_{i}^{(j)}|\boldsymbol{s}^{(j)})\sum_{a_{-i}^{(j-1)}}\boldsymbol{P}(\boldsymbol{s}^{(j)}|\boldsymbol{s}^{(j-1)},\boldsymbol{a}^{(j-1)})\pi_{-i}'(a_{-i}^{(j)}|\boldsymbol{s}^{(j)}).
  \end{aligned}
\end{equation}

Then if the multipliers are the probability of Eq. \eqref{mtl} of corresponding sequences, the dual function can be expressed as
\begin{equation}
  \begin{aligned}
   \lim_{t\to \infty} L_{\boldsymbol{r}}^{(t+1)}(\boldsymbol{\pi}',\lambda_{\boldsymbol{\pi}}') &= \sum\limits_{i=1}^N\mathbb{E}_{\pi_i,\pi_{-i}'}[r_i(\boldsymbol{s},\boldsymbol{a})]\\
  & -\sum\limits_{i=1}^N\mathbb{E}_{\pi_i'\pi_{-i}'}[r_i(\boldsymbol{s},\boldsymbol{a})].
\end{aligned}
\end{equation}

\textbf{\textit{Proof 3:}} Here we denote that $Q_i'(\tau_i) = Q_i(\tau_i;\boldsymbol{\pi}',\boldsymbol{r})$,  $\hat{q}'$ and $\hat{v}'$ are defined similarly. According to Eq. \eqref{ldf}
\begin{equation}
  \label{sdf}
  L_{\boldsymbol{r}}^{(t+1)}(\boldsymbol{\pi}',\lambda_{\boldsymbol{\pi}}')=\sum_{i=1}^N\sum_{\tau_i\in\mathcal{T}_i}\lambda'(\tau_i)(Q_i'(\tau_i)-\hat{v}_i'(\boldsymbol{s}^{(0)})).
\end{equation}

We expand the Eq. \eqref{sdf} by the definition of $Q_i'$ and $\hat{v}_i'$, it can be noticed that
\begin{equation}
  \label{cvg}
  \begin{aligned}
  &\sum_{\tau_i\in\mathcal{T}_i}\lambda'(\tau_i)Q_i'(\tau_i)\\
  &=\mathbb{E}_{\pi_i}[\mathbb{E}_{\pi_{-i}'}[\sum_{j=0}^{t-1}\gamma^jr_i(\boldsymbol{s}^{(j)},\boldsymbol{a}^{(j)})+\gamma^t\hat{q}_i'(\boldsymbol{s}^{(t)},a_i^{(t)})]],
  \end{aligned}
\end{equation}
which means that $\pi_i$ is used for executing the first $t$ timesteps, while $\pi_i'$ is used subsequently, with other agents complying $\pi_{-i}'$ all along. When $t \to \infty$, $\gamma^t \to 0$ and $\hat{q}_i'(\boldsymbol{s}^{(t)},a_i^{(t)})$ is bounded, the Eq. \eqref{cvg} converges to $\mathbb{E}_{\pi_i,\pi_{-i}'}[r_i]$. Then, for the term $\hat{v}_i'(\boldsymbol{s}^{(0)})$, it can be observed that
\begin{equation}
  \label{48}
  \sum_{\tau_i\in\mathcal{T}_i}\lambda'(\tau_i)\hat{v}_i'(\boldsymbol{s}^{(0)}) = \mathbb{E}_{\boldsymbol{s}^{(0)} \sim \boldsymbol{P}^{(0)}(\boldsymbol{s})}[\hat{v}_i'(\boldsymbol{s}^{(0)})] = \mathbb{E}_{\boldsymbol{\pi}'}[r_i].
\end{equation}

Combining Eq. \eqref{cvg} and Eq. \eqref{48}, the Theorem 2 can be proved. $\hfill\blacksquare$

Based on analysis before, we define the MAIRL procedure similar to Eq. \eqref{gail2}
\begin{equation}
  \label{mairl}
  \begin{aligned}
     \mathrm{MAIRL}_{\psi}&(\boldsymbol{\pi}_{E})=\mathop{\rm{argmin}}\limits_{\mathbf{r}}-\psi(\mathbf{r})+\sum_{i=1}^{N}(\mathbb{E}_{\boldsymbol{\pi}_{E}}[r_{i}])\\
    &\quad\quad  -\max_{\boldsymbol{\pi}}\sum_{i=1}^{N}(\beta H_{i}(\pi_{i})+\mathbb{E}_{\pi_{i},\pi_{E_{-i}}}[r_{i}]),
  \end{aligned}
\end{equation}
where $H_i(\pi_i)$ is the casual entropy of $\pi_i$ and $\beta$ is the hyper parameter represents the strength of regularization. If $N=1$ and $\beta=1$, the Eq. \eqref{mairl} is equivalent to the Eq. \eqref{gail2}. 

Finally, we can derive the solution of the dual problem:

\textbf{\textit{Theorem 3:}} Assume that the reward regularizer is additively separable for each AUV, namely $\phi(\boldsymbol{r}) = \sum^{N}_{i=1}\phi_i(r_i)$, and for all feasible $r \in \text{MAIRL}_\phi(\pi_E)$ there is a unique solution for MARL($r$). Then, the dual optimum can be expressed as
\begin{equation}
  \label{fin}
  \begin{aligned}
  &\; \text{MARL} \circ \text{MAIRL}_\psi(\boldsymbol{\pi}_E) \\ 
  &= \mathop{\rm{argmin}}\limits_{\pi \in \Pi}\sum^{N}_{i=1}-\beta H_i(\pi_i)+\psi_i^{\star}(\rho_{\pi_{i},\pi_{E_{-i}}}-\rho_{\boldsymbol{\pi}_{E}}).
  \end{aligned}
\end{equation}

\textbf{\textit{Proof 4:}} We attempt to use Eq. \eqref{gail3} to solve the dual problem by decomposing the problem to the single-agent scenario. The RL objective for $i$-th AUV, where other AUVs complies policy $\pi_{E_{-i}}$, can be expressed as
\begin{equation}
  \text{RL}_i(r_i)=\mathop{\rm{max}}\limits_{\pi_i \in \Pi} H_i(\pi_i)+\mathbb{E}_{\pi_{i},\pi_{E_{-i}}}[r_i].
\end{equation}

Then the IRL objective of the same condition can be expressed as
\begin{equation}
  \begin{aligned}
   \text{IRL}_{i,\psi}(\boldsymbol{\pi}_E) &= \mathop{\rm{argmin}}\limits_{r_i \in \mathbb{R}^{\boldsymbol{S}_i \times \boldsymbol{A}_i}} -\psi_i(r_i)+\mathbb{E}_{\pi_{E_i}}[r_i] \\
  & -\left(\mathop{\rm{max}}\limits_{\pi_i \in \Pi}H_i(\pi_i) + \mathbb{E}_{\pi_i,\pi_{E_{-i}}}[r_i]\right).
  \end{aligned}
\end{equation}

As we have assumed that the reward regularizer is additively separable, the solution to MAIRL can be expressed as a group of solutions of IRL
\begin{equation}
  \text{MAIRL}_\psi = [\text{IRL}_{1,\psi},\ldots,\text{IRL}_{N,\psi}].
\end{equation}

Similarly, 
\begin{equation}
  \text{MARL}_{\boldsymbol{r}} = [\text{RL}_1(r_1),\ldots,\text{RL}_N(r_N)].
\end{equation}

Then, we can solve the dual problem for each AUV analogous to Eq. \eqref{gail3}, and Theorem 3 can be proved.$\hfill\blacksquare$

Analogous to Eq. \eqref{gail4}, we can design a regularizer like $\psi_{\text{GA}}$ in single-agent scenario. As the optimum solution for $\psi^\star_{\text{GA}}(\rho_{\pi_i,\pi_{E_{-i}}},\rho_{\boldsymbol{\pi}_E})$ is same to $\psi^\star_{\text{GA}}(\rho_{\boldsymbol{\pi}},\rho_{\boldsymbol{\pi}_E})$, namely $\boldsymbol{\pi}_E$, we can substitute the former of in Eq. \eqref{fin} with the latter, and eventually the Eq. \eqref{wtproof} can be derived. Specifically, the decentralized setting of the discriminator utilizes the regularizer of $\psi_i(r_i) = \psi_\text{GA}(r_i)$, and the centralized setting utilizes the regularizer as follows:
 \begin{equation}
  \psi(\boldsymbol{r}) = \begin{cases}
    \psi_\text{GA}(\boldsymbol{r}), & \text{if} \quad r_1=\cdots=r_N, \\
    \infty, & \text{otherwise}.
  \end{cases}
 \end{equation}

% \textit{Proposition 1:} Given that $\beta=0$,
% \begin{equation}
%   g(x)=\left\{\begin{array}{ll}-x-\log(1-e^x),&if\quad r_i>0\\+\infty,&otherwise\end{array}\right.
% \end{equation}

% \begin{equation}
%   \arg\min\sum_{i=1}^N\psi_i^\star(\rho_{\pi_i,\pi_{E_{-i}}}-\rho_{\pi_E})=\arg\min_\pi\sum_{i=1}^N\psi_i^\star(\rho_{\pi_i,\pi_{-i}}-\rho_{\pi_E})=\pi_E
% \end{equation}
% \textit{Proof 5:} 



%\begin{thebibliography}{1}
\bibliographystyle{IEEEtran}
\bibliography{mybibliography}

%\end{thebibliography}

% \newpage % optional page break

 

% \end{IEEEbiography}



\begin{IEEEbiography}[{\includegraphics[width=1in,height=1.25in,clip,keepaspectratio]{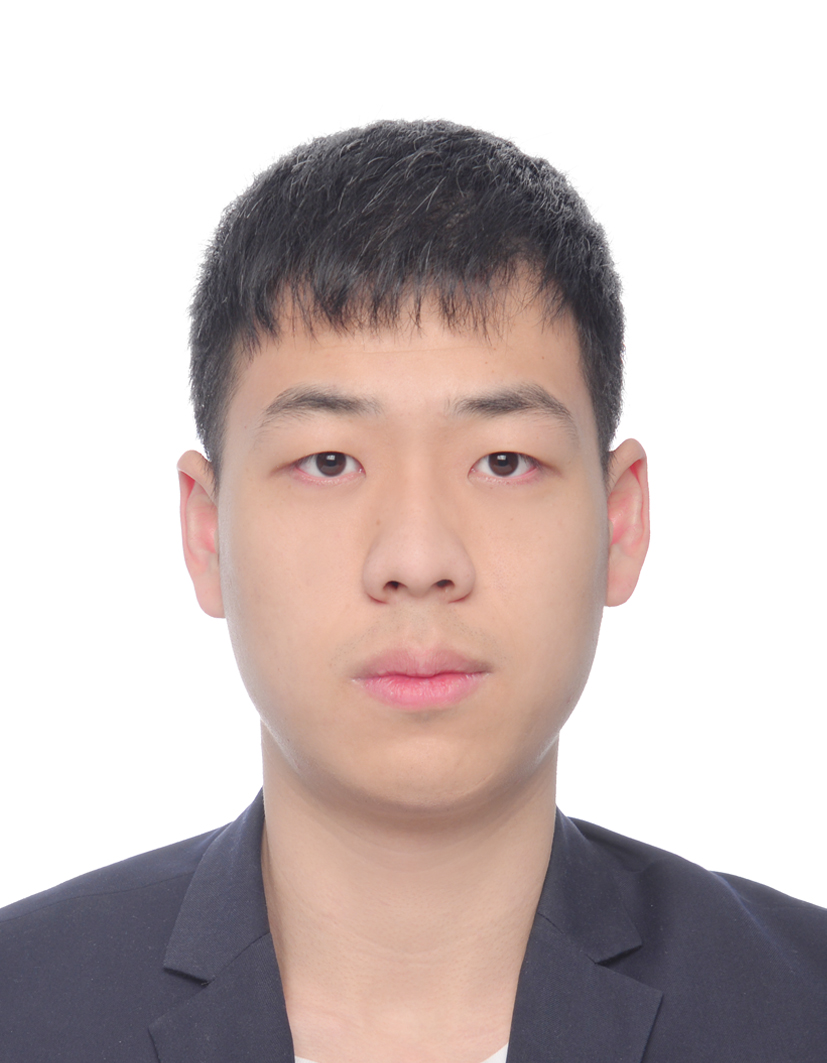}}]{Guanwen Xie} (Student Member, IEEE) received his B.E. degree in Ocean Engineering and Technology at Ocean College from Zhejiang University, and he is currently pursuing the M.S. degree in Electronic Information from Tsinghua Shenzhen International Graduate School, Tsinghua University, China. His main research interest is applying reinforcement learning and large language models to underwater intelligent decision-making. Besides, he is also an outstanding graduate of Zhejiang University.
\end{IEEEbiography}

\begin{IEEEbiography}
[{\includegraphics[width=1in,height=1.25in,clip,keepaspectratio]{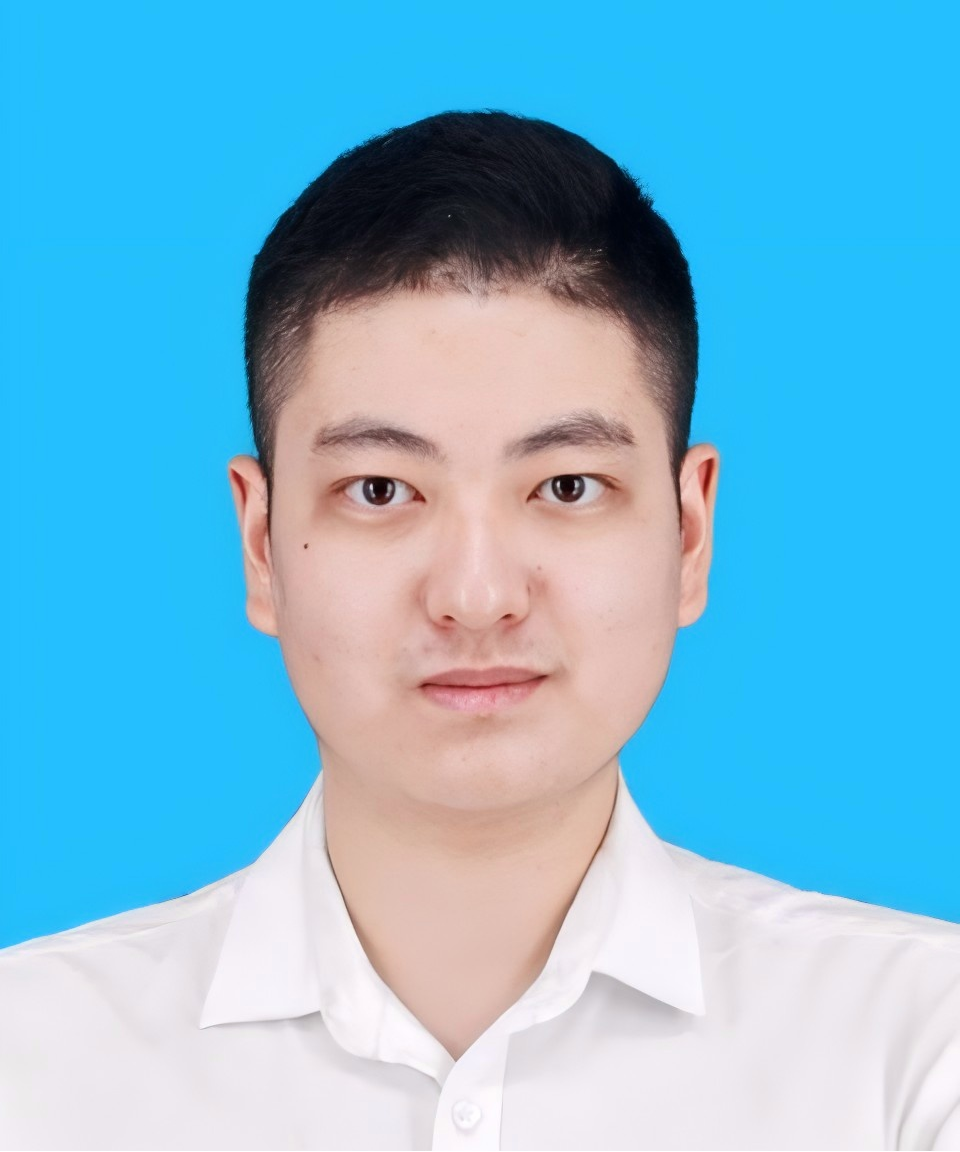}}]{Jingzehua Xu} (Student Member, IEEE)
received his B.S. degree in Marine Science and B.E. degree in Electronic Science and Technology from Zhejiang University, Hangzhou, China in 2023. He also received his MicroM.S. certificate in Statistics and Data Science from the Massachusetts Institute of Technology (MIT), Boston, USA in 2025. He is currently pursuing the M.S. degree in Electronic and Information Engineering from Tsinghua Shenzhen International Graduate School, Tsinghua University, China. He is also a guest student at Woods Hole Oceanographic Institution (WHOI) and will be a research assistant at the University of Hong Kong (HKU). His main research interests include reinforcement learning, large language models, and their applications in planning and control. Besides, he is the outstanding graduate at Zhejiang University.
\end{IEEEbiography}

\begin{IEEEbiography}
[{\includegraphics[width=1in,height=1.25in,clip,keepaspectratio]{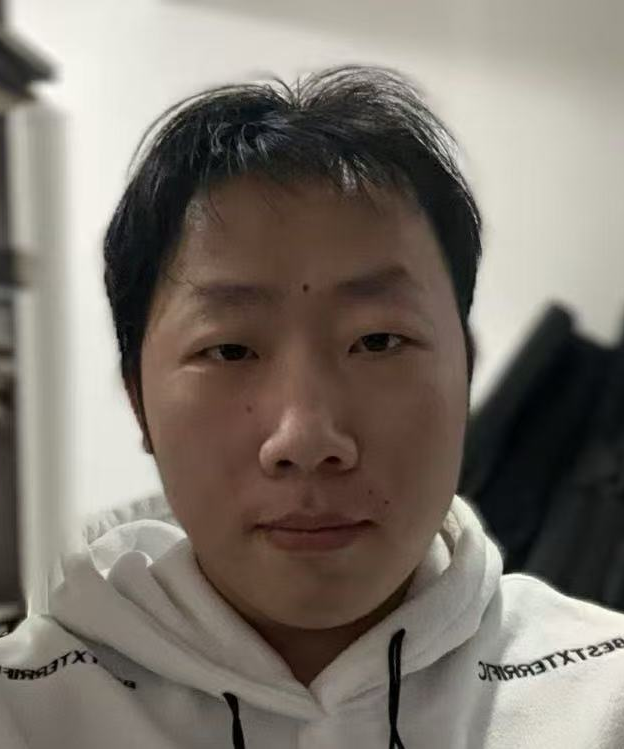}}]{Ziqi Zhang} is currently a part-time research consulant in WorldQuant. He served as a research assistant at School of Life Science, Tsinghua University, Beijing, China in 2021. From 2022 to 2024, he served as a research assistant at School of Engineering, Westlake University, under the supervision of Prof. Donglin Wang. His main research interests include reinforcement learning, and large language models.
\end{IEEEbiography}

\begin{IEEEbiography}[{\includegraphics[width=1.1in,height=1.33in]{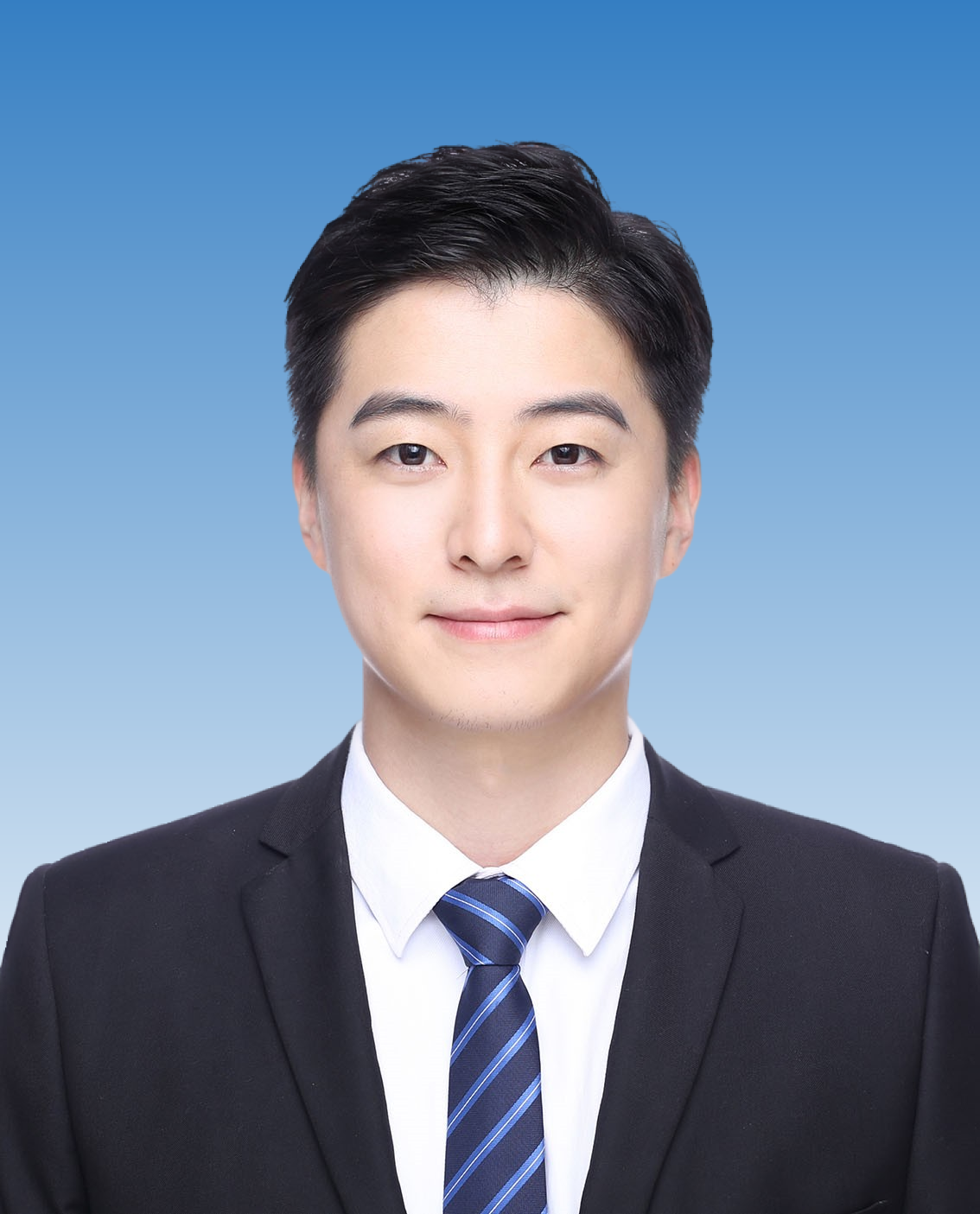}}]{\textbf{Xiangwang Hou}} (Member, IEEE) is currently a postdoctoral researcher in the Department of Electronic Engineering, Tsinghua University, Beijing, China. He received his B.S. degree from Shandong University of Technology in 2017, his M.S. degree from Xidian University in 2020, and his Ph.D. degree from Tsinghua University in 2025. From 2023 to 2024, he was a Joint Ph.D. student at the School of Computer Science and Engineering, Nanyang Technological University, Singapore, under the supervision of Prof. Dusit Niyato. From 2020 to 2021, he worked as an algorithm engineer at Huawei Technologies Co., Ltd. and at Tsinghua University. His research interests include edge intelligence, federated learning, wireless AI, and UAV/AUV networks. He was a recipient of the Best Paper Award of IEEE ICC.
\end{IEEEbiography}

\begin{IEEEbiography}[{\includegraphics[width=1.1in,height=1.33in]{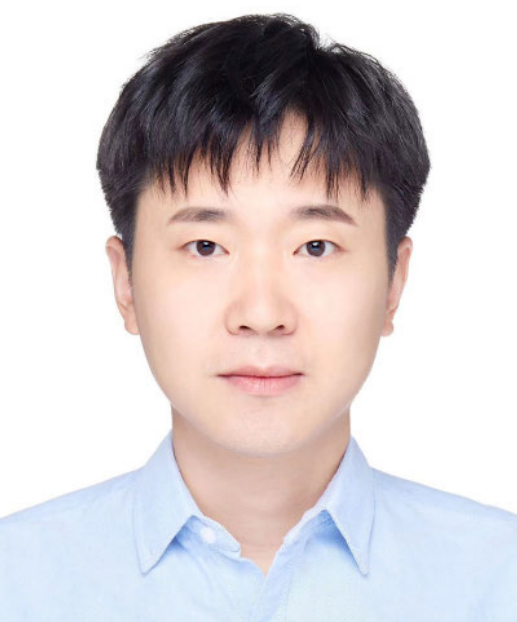}}]{\textbf{Dongfang Ma}} (Member, IEEE) received the M.S. and Ph.D. degrees in information engineering and control from Jilin University, Changchun, China, in 2009 and 2012, respectively. From 2012 to 2014, he was a Post-Doctoral Researcher with the Department of Civil Engineering and Architecture, Zhejiang University, where he has been with the Institute of Marine Sensing and Networking since 2015. Since 2019, he has also been a part-time Research Fellow with the Peng Cheng Laboratory. His current research interests include transportation big data mining and intelligent transportation systems.
 \end{IEEEbiography}

\begin{IEEEbiography}[{\includegraphics[width=1in,height=1.25in,clip,keepaspectratio]{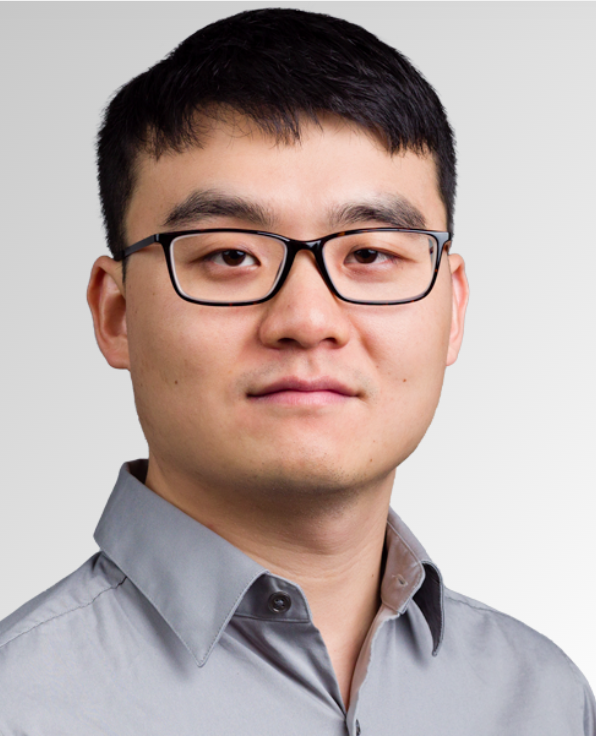}}]{Shuai Zhang} 
(Member, IEEE) received his B.E. degree from the University of Science and Technology of China, Hefei, China, in 2016, and his Ph.D. degree from Rensselaer Polytechnic Institute, Troy, NY, USA, in 2021. From 2022 to 2023, he was a Postdoctoral Research Associate at Rensselaer Polytechnic Institute. He is currently an Assistant Professor in the Department of Data Science at the New Jersey Institute of Technology, Newark, NJ, USA. His research focuses on the theoretical foundations of deep learning and the development of principled, efficient algorithms to improve the reliability and performance of AI applications. He has served as a reviewer or program committee member for NeurIPS, ICML, AAAI, ICLR, AISTATS, TMLR, IEEE TSP, IEEE TNNLS, and IEEE TIT.
\end{IEEEbiography}

\begin{IEEEbiography}[{\includegraphics[width=1in,height=1.25in,clip,keepaspectratio]{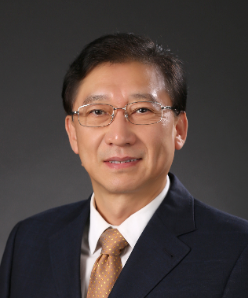}}]{Yong Ren} 
(Senior Member, IEEE) is currently a Professor with the Department of Electronics Engineering and the Director of the Complexity Engineered Systems Lab. He received the B.S.,
M.S., and Ph.D. degrees in electronic engineering from Harbin Institute of Technology, Harbin, China, in 1984, 1987, and 1994, respectively. He worked as a Postdoctoral Fellow with the Department of Electronics Engineering, Tsinghua University, Beijing, China, from 1995 to 1997. He holds 60 patents and has authored or co-authored more than 300 technical papers in communication and signal processing. His current research interests include maritime information networks and swarm intelligence.
\end{IEEEbiography}

\begin{IEEEbiography}[{\includegraphics[width=1in,height=1.25in,clip,keepaspectratio]{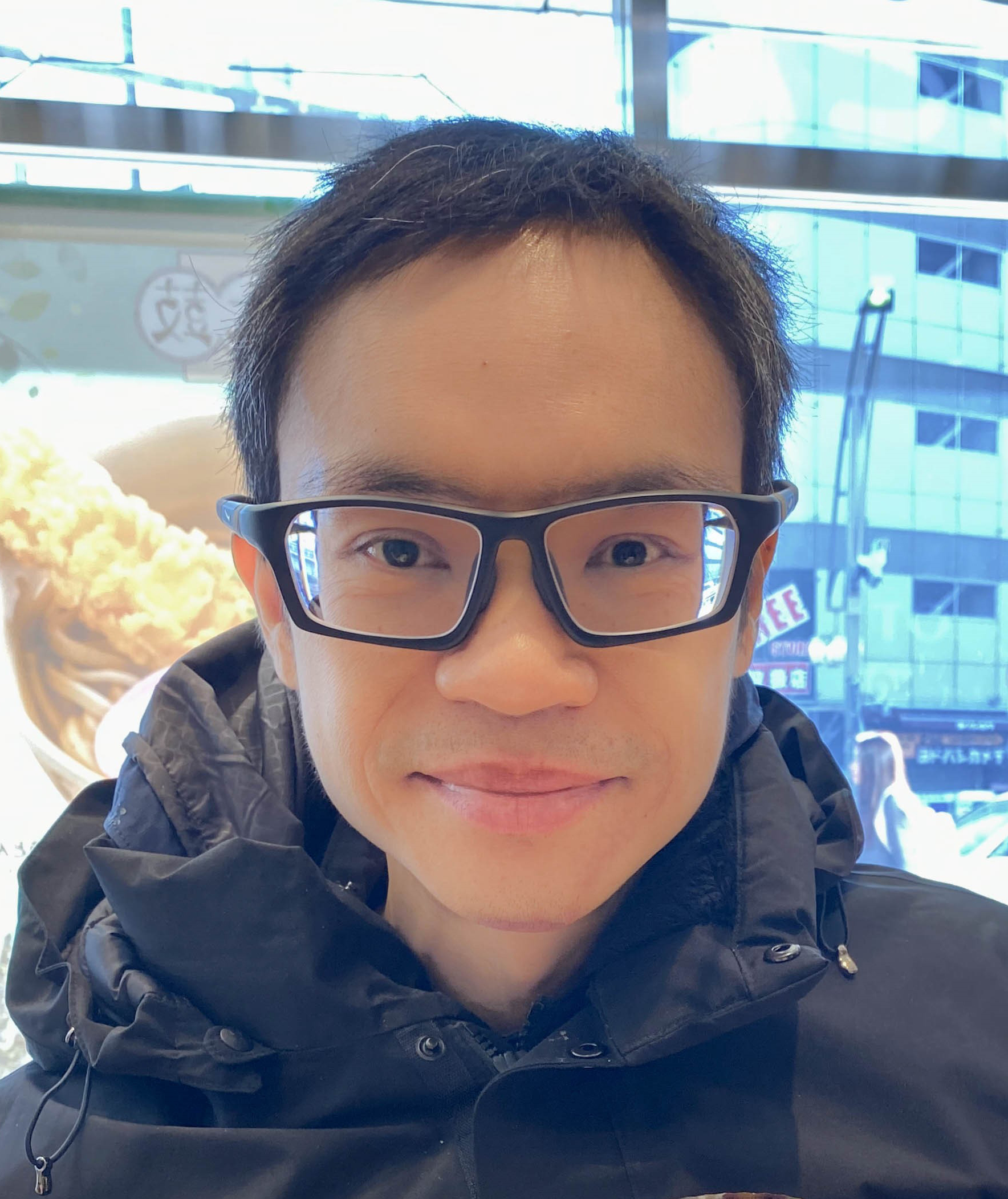}}]{Dusit Niyato} (Fellow, IEEE) is a professor in the College of Computing and Data Science, at Nanyang Technological University, Singapore. He received B.Eng. from King Mongkuts Institute of Technology Ladkrabang (KMITL), Thailand and Ph.D. in Electrical and Computer Engineering from the University of Manitoba, Canada. His research interests are in the areas of mobile generative AI, edge general intelligence, quantum computing and networking, and incentive mechanism design.
\end{IEEEbiography}

% \vfill

\end{document}